\documentclass{article}
\usepackage{graphicx} % Required for inserting images
\usepackage{amssymb
}

\usepackage{enumitem}
\usepackage{tabularx}
\usepackage{booktabs}
\usepackage{float}
\usepackage{caption}
\usepackage{microtype}

\newcommand{\Expect}{\bf\rm \mathbb{E}}

\begin{document}

%standard macros for typesetting articles
\hyphenation{super-terse mea-sure semi-recursive non-recursive
             non-superterse}
\newcounter{savenumi}
\newenvironment{savenumerate}{\begin{enumerate}
\setcounter{enumi}{\value{savenumi}}}{\end{enumerate}
\setcounter{savenumi}{\value{enumi}}}
\newtheorem{theoremfoo}{Theorem}%[section] %by chapter in report style
\newenvironment{theorem}{\pagebreak[1]\begin{theoremfoo}}{\end{theoremfoo}}
\newenvironment{repeatedtheorem}[1]{\vskip 6pt
\noindent
{\bf Theorem #1}\ \em
}{}
%\newtheorem{repeatedtheorem}{Theorem} % number this yourself using
%                                      % \def\therepeatedtheorem{text}
%\newtheorem{observationfoo}[theoremfoo]{Observation}
%\newenvironment{observation}{\pagebreak[1]\begin{observationfoo}}{\end{observationfoo}}
\newtheorem{propositionfoo}[theoremfoo]{Proposition}
\newenvironment{proposition}{\pagebreak[1]\begin{propositionfoo}}{\end{propositionfoo}}
\newtheorem{lemmafoo}[theoremfoo]{Lemma}
\newenvironment{lemma}{\pagebreak[1]\begin{lemmafoo}}{\end{lemmafoo}}
\newtheorem{conjecturefoo}[theoremfoo]{Conjecture}
\newenvironment{conjecture}{\pagebreak[1]\begin{conjecturefoo}}{\end{conjecturefoo}}
\newtheorem{corollaryfoo}[theoremfoo]{Corollary}
\newenvironment{corollary}{\pagebreak[1]\begin{corollaryfoo}}{\end{corollaryfoo}}
\newtheorem{exercisefoo}{Exercise}
\newenvironment{exercise}{\pagebreak[1]\begin{exercisefoo}\rm}{\end{exercisefoo}}
\newtheorem{openfoo}[theoremfoo]{Question}
\newenvironment{open}{\pagebreak[1]\begin{openfoo}}{\end{openfoo}}
\newtheorem{nttn}[theoremfoo]{Notation}
\newenvironment{notation}{\pagebreak[1]\begin{nttn}\rm}{\end{nttn}}

\newtheorem{dfntn}[theoremfoo]{Definition}
\newenvironment{definition}{\pagebreak[1]\begin{dfntn}\rm}{\end{dfntn}}

\newenvironment{proof}
    {\pagebreak[1]{\narrower\noindent {\bf Proof:\quad\nopagebreak}}}{\QED}
%\newenvironment{alternateproof}
%               {{\narrower\noindent {\bf Alternate proof:\quad}}}{\QED}
%\newenvironment{sketch}
%    {\pagebreak[1]{\narrower\noindent {\bf Proof sketch:\quad\nopagebreak}}}{}
\newenvironment{sketch}
    {\pagebreak[1]{\narrower\noindent {\bf Proof sketch:\quad\nopagebreak}}}{\QED}
\newenvironment{comment}{\penalty -50 $(*$\nolinebreak\ }{\nolinebreak $*)$\linebreak[1]\ }
\renewcommand{\theenumi}{\roman{enumi}}
 % number subtheorems i., ii., iii., iv.,...
\newcommand{\yyskip}{\penalty-50\vskip 5pt plus 3pt minus 2pt}
\newcommand{\blackslug}{\hbox{\hskip 1pt
        \vrule width 4pt height 8pt depth 1.5pt\hskip 1pt}}

\newcommand{\dash}{{\rm\mbox{-}}}
\newcommand{\floor}[1]{\left\lfloor#1\right\rfloor}
\newcommand{\ceiling}[1]{\left\lceil#1\right\rceil}
\newcommand{\st}{\mathrel{:}}
\newcommand{\ang}[1]{\langle#1\rangle}
\newcommand{\DTIME}{{\rm DTIME}}
\newcommand{\DSPACE}{{\rm DSPACE}}
\newcommand{\NSPACE}{{\rm NSPACE}}
\newcommand{\polylog}{\mathop{\rm polylog}}
\newcommand{\UP}{{\rm UP}}
\newcommand{\PH}{{\rm PH}}
\newcommand{\R}{{\rm R}}
\newcommand{\NTIME}{{\rm NTIME}}
\newcommand{\PTIME}{{\rm PrTIME}}
\newcommand{\PSPACE}{{\rm PSPACE}}
\newcommand{\NE}{{\rm NE}}
\newcommand{\poly}{{\rm poly}}
\newcommand{\LOGSPACE}{{\rm LOGSPACE}}
\newcommand{\E}{{\rm E}}
\newcommand{\SAT}{{\rm SAT}}
\newcommand{\NP}{{\rm NP}}
\let\paragraphsym\P
\renewcommand{\P}{{\rm P}}      % used to be paragraph sign
\newcommand{\coNP}{{\co\NP}}
\newcommand{\propersubset}{\subset}
\newcommand{\xor}{\oplus}
\newcommand{\m}{{\rm m}}
\newcommand{\T}{{\rm T}}
\let\savett\tt
\renewcommand{\tt}{{\rm tt}}    % used to be typewriter type style
\newcommand{\btt}{{\rm btt}}
\newcommand{\ntt}{{n\rm\mbox{-}tt}}
\newcommand{\ktt}{{k\rm\mbox{-}tt}}
\newcommand{\kT}{{k\rm\mbox{-}T}}
\newcommand{\onett}{{1\rm\mbox{-}tt}}
\newcommand{\Abar}{{\bar{A}}}
\newcommand{\Bbar}{{\bar{B}}}
\newcommand{\Cbar}{{\bar{C}}}
\def\nre.{$n$\/-r.e.}
\newcommand{\ie}{{\it i.e.}}
\newcommand{\eg}{{\it e.g.}}
\newenvironment{multi}{\left\{\begin{array}{ll}}{\end{array}\right.}
\newcommand{\union}{\cup}
\newcommand{\Union}{\bigcup}
\newcommand{\intersection}{\cap}
\newcommand{\Intersection}{\bigcap}
\newcommand{\OR}{\vee}
\newcommand{\AND}{\wedge}
\newcommand{\reason}[1]{\mbox{\qquad\rm #1}}     % added quad 8/9, qquad 8/31
\newcommand{\nlreason}[1]{\\ & & \mbox{\rm #1}}  % killed comma 8/31
\newcommand{\infinity}{\infty}
\newcommand{\Implies}{\mathrel{\Longrightarrow}}
\newcommand{\Iff}{\mathrel{\Longleftrightarrow}}
\newcommand{\scrod}{\quad\nopagebreak}
\newcommand{\ACCEPT}{{\rm ACCEPT}}
\newcommand{\PARITY}{{\rm PARITY}}
\newcommand{\PARITYP}{\PARITY\P}
\newcommand{\MOD}{{\rm MOD}}
\newcommand{\PP}{{\rm PP}}
\newcommand{\FewP}{{\rm FewP}}
\newcommand{\EP}{{\rm EP}}
\tolerance=2000
\def\pmod#1{\allowbreak\mkern6mu({\rm mod}\,\,#1)}
\newcommand{\co}{{\rm co}\dash}
\newcommand{\PF}{{\rm PF}}
\newcommand{\EXP}{{\rm EXP}}
\newcommand{\NEXP}{{\rm NEXP}}
\newcommand{\Hastad}{H{\aa}stad}
\newcommand{\Kobler}{K\"obler}
\newcommand{\Schoning}{Sch\"oning}
\newcommand{\Toran}{Tor\'an}
\newcommand{\Balcazar}{Balc{\'a}zar}
\newcommand{\Diaz}{D\'{\i}az}
\newcommand{\Gabarro}{Gabarr{\'o}}
\newcommand{\RIA}{CCR-8808949}
\newcommand{\PYI}{CCR-8958528}
\newcommand{\NSFFT}{CCR-9415410}
\newcommand{\NSFALEXIS}{CCR-9522084}
\newtheorem{factfoo}[theoremfoo]{Fact}
\newenvironment{fact}{\pagebreak[1]\begin{factfoo}}{\end{factfoo}}
\newenvironment{acknowledgments}{\par\vskip 20pt\noindent{\footnotesize\em Acknowledgments.}\footnotesize}{\par}
\newcommand{\POLYLOGSPACE}{{\rm POLYLOGSPACE}}
\newcommand{\SPACE}{{\rm SPACE}}
\newcommand{\BPP}{{\rm BPP}}
\newcommand{\lookatme}{\marginpar[$\Rightarrow$]{$\Leftarrow$}}
\newcommand{\muchless}{\ll}
\newcommand{\plusminus}{\pm}
\newcommand{\nopagenumbers}{\renewcommand{\thepage}{\null}}
\newenvironment{algorithm}{\renewcommand{\theenumii}{\arabic{enumii}}\renewcommand{\labelenumii}{Step \theenumii :}\begin{enumerate}}{\end{enumerate}}
\newcommand{\squeeze}{
\textwidth 6in
\textheight 8.8in
\oddsidemargin 0.2in
\topmargin -0.4in
}
\newcommand{\existsunique}{\exists \! ! \,}
\newcommand{\YaleAddress}{Yale University,
Dept.\ of Computer Science,
P.O.\ Box 208285, Yale Station,
New Haven, CT\ \ 06520-8285.
Email: {\savett beigel-richard@cs.yale.edu}}
\newcommand{\pageline}[2]{[{\bf p.~#1, l.~#2}]}
\newcommand{\Beigel}{Richard Beigel\thanks{\newsupport} \\
Yale University \\
Dept.\ of Computer Science \\
P.O. Box 208285, Yale Station \\
New Haven, CT\ \ 06520-8285}

\newcommand{\newsupport}{Supported in part by grants \RIA\ and \PYI\
from the National Science Foundation}
\newtheorem{propertyfoo}[theoremfoo]{Property}
\newenvironment{property}{\pagebreak[1]\begin{propertyfoo}}{\end{propertyfoo}}

\makeatletter

\def\@makechapterhead#1{ \vspace*{50pt} { \parindent 0pt \raggedright 
 \ifnum \c@secnumdepth >\m@ne \huge\bf \@chapapp{} \thechapter. \par 
 \vskip 20pt \fi \Huge \bf #1\par 
 \nobreak \vskip 40pt } }

\def\@sect#1#2#3#4#5#6[#7]#8{\ifnum #2>\c@secnumdepth
     \def\@svsec{}\else 
     \refstepcounter{#1}\edef\@svsec{\csname the#1\endcsname.\hskip 1em }\fi
     \@tempskipa #5\relax
      \ifdim \@tempskipa>\z@ 
        \begingroup #6\relax
          \@hangfrom{\hskip #3\relax\@svsec}{\interlinepenalty \@M #8\par}
        \endgroup
       \csname #1mark\endcsname{#7}\addcontentsline
         {toc}{#1}{\ifnum #2>\c@secnumdepth \else
                      \protect\numberline{\csname the#1\endcsname}\fi
                    #7}\else
        \def\@svsechd{#6\hskip #3\@svsec #8\csname #1mark\endcsname
                      {#7}\addcontentsline
                           {toc}{#1}{\ifnum #2>\c@secnumdepth \else
                             \protect\numberline{\csname the#1\endcsname}\fi
                       #7}}\fi
     \@xsect{#5}}

\def\@begintheorem#1#2{\it \trivlist \item[\hskip \labelsep{\bf #1\ #2.}]}

\def\@opargbegintheorem#1#2#3{\it \trivlist
      \item[\hskip \labelsep{\bf #1\ #2\ (#3).}]}

\makeatother

%%Clean up section headers.  Period after section number.  Smaller font.
%\makeatletter
%\def\section{\@startsection {section}{1}{\z@}{-3.5ex plus -1ex minus 
% -.2ex}{2.3ex plus .2ex}{\large\bf}}
%\makeatother
%\def\thesection {\arabic{section}.}
%\def\thesubsection {\thesection\arabic{subsection}.}
%\def\thesubsubsection {\thesubsection\arabic{subsubsection}.}
%\def\theparagraph {\thesubsubsection\arabic{paragraph}.}
%\def\thesubparagraph {\theparagraph\arabic{subparagraph}.}

\newcommand{\kstar}{{\textstyle *}}
\newcommand{\F}[2]{F_{#1}^{#2}}
\newcommand{\FQ}[2]{\PF_{{#1\rm\mbox{-}T}}^{#2}}
\newcommand{\Q}[2]{\P_{{#1\rm\mbox{-}T}}^{#2}}
\newcommand{\FQp}[2]{\PF_{{#1\rm\mbox{-}tt}}^{#2}}
\newcommand{\Qp}[2]{\P_{{#1\rm\mbox{-}tt}}^{#2}}
\newcommand{\ppoly}{\P/\poly}
\newcommand{\pnplog}{\P^{\NP[\log]}}
\newcommand{\pleq}[1]{\leq_{#1}^{p}}

\def\abs#1{\left|#1\right|}
\def\exp#1{{\rm exp}\left(#1\right)}
\newcommand{\transfig}[1]{\begin{center}\vspace{-6pt}\input{#1.tex}\vspace{-12pt}\end{center}}

%%%%%%%%%%%%%%%%%%%%%%%%%%%%%%%%%%%%%%%%%%%%%%%%%%%%%%%%%%%%%%%%
%%%%%%%%%%%%%%%%%%%%%% Bibliography Stuff %%%%%%%%%%%%%%%%%%%%%%
%%%%%%%%%%%%%%%% By Richard Beigel and Nick Reingold %%%%%%%%%%%
%%%%%%%%%%%%%%%%%%%%%%%%%%%%%%%%%%%%%%%%%%%%%%%%%%%%%%%%%%%%%%%%

\newif\ifshortconferences
\shortconferencesfalse
\newif\ifmediumconferences
\mediumconferencesfalse

\def\ending#1{{\count1=#1\relax
% mod out by 100 first
\count2=\count1
\divide\count2 by 100
\multiply\count2 by 100
\advance\count1 by -\count2
\ifnum\count1=11
th%
\else \ifnum\count1=12
th%
\else \ifnum\count1=13
th%
\else 
\count2=\count1
\divide\count1 by 10
\multiply\count1 by 10
\advance\count2 by -\count1
\ifnum\count2=1
st%
\else \ifnum\count2=2
nd%
\else \ifnum\count2=3
rd%
\else th%
\fi\fi\fi\fi\fi\fi
}}

\def\STOC{\conf{STOC}}
\def\STOCname{\ifshortconferences ACM STOC\else\ifmediumconferences Ann. ACM Symp. Theor. Comput.\else Annual ACM Symposium on Theory of Computing\fi\fi}
\def\STOCzero{68}

\def\FOCS{\conf{FOCS}}
\def\FOCSname{\ifshortconferences IEEE FOCS\else\ifmediumconferences Ann. Symp. Found. Comput. Sci.\else IEEE Symposium on Foundations of Computer Science\fi\fi}
\def\FOCSzero{59}

\def\FSTTCS{\conf{FSTTCS}}
\def\FSTTCSname{\ifshortconferences FST\&TCS\else\ifmediumconferences Ann. Conf. Found. Softw. Theor. and Theor. Comp. Sci.\else Conference on Foundations of Software Theory and Theoretical Computer Science\fi\fi}
\def\FSTTCSzero{80}

\def\Complexity{\conf{Complexity}}
\def\Complexityname{\ifshortconferences Conf. Computational Complexity\else\ifmediumconferences Ann. Conf. Computational Complexity\else Annual Conference on Computational Complexity\fi\fi}
\def\Complexityzero{85}
\def\ComplexityOne{Structure in Complexity Theory}

\def\Structures{\conf{Structures}}
\def\Structuresname{\ifshortconferences Ann. Conf. Structure in Complexity Theory\else\ifmediumconferences Ann. Conf. Structure in Complexity Theory\else Annual Conference on Structure in Complexity Theory\fi\fi}
\def\Structureszero{85}
\def\StructuresOne{Structure in Complexity Theory}

\def\SODA{\conf{SODA}}
\def\SODAname{\ifshortconferences ACM-SIAM Symp. on Discrete Algorithms\else Annual ACM-SIAM Symposium on Discrete Algorithms\fi}
\def\SODAzero{89}

\def\STACS{\conf{STACS}}
\def\STACSname{\ifshortconferences STACS\else\ifmediumconferences\else Annual Symposium on Theoretical Aspects of Computer Science\fi\fi}
\def\STACSzero{83}

\def\SPAA{\conf{SPAA}}
\def\SPAAname{\ifshortconferences ACM SPAA\else\ifmediumconferences Ann. ACM Symp. Par. Alg. Arch.\else Annual ACM Symposium on Parallel Algorithms and Architectures\fi\fi}
\def\SPAAzero{88}

\def\Proceedings{\ifshortconferences Proc.\else\ifmediumconferences Proc.\else Proceedings\fi\fi}
\def\Proceedingsofthe{\ifshortconferences Proc.\else\ifmediumconferences Proc.\else Proceedings of the\fi\fi}

\newcounter{confnum}

\def\conf#1#2{%
\setcounter{confnum}{#2}%
\addtocounter{confnum}{-\csname #1zero\endcsname}%
\ifnum\value{confnum}=1%
\expandafter\ifx\csname #1One\endcsname\relax%
\Proceedingsofthe\ \arabic{confnum}\ending{\value{confnum}}\ \csname #1name\endcsname%
\else \csname #1One\endcsname\fi%
\else%
\Proceedingsofthe\
\arabic{confnum}\ending{\value{confnum}}\ \csname #1name\endcsname\fi}

\def\qsym{\vrule width0.7ex height0.9em depth0ex}

%%
%% A switch to turn the symbol on or off.  Normally it is on.
%%
\newif\ifqed\qedtrue

%%
%% The next invocation of \qed in suppressed with this
%% command.
%% 
\def\noqed{\global\qedfalse}

%%
%% This definition is (essentially) from the TeXbook.
%%
\def\qed{\ifqed{\penalty1000\unskip\nobreak\hfil\penalty50
\hskip2em\hbox{}\nobreak\hfil\qsym
\parfillskip=0pt \finalhyphendemerits=0\par\medskip}\fi\global\qedtrue}

\def\QEDcomment#1{\ifqed{\penalty1000\unskip\nobreak\hfil\penalty50
\hskip2em\hbox{}\nobreak\hfil\qsym\ #1
\parfillskip=0pt \finalhyphendemerits=0\par\medskip}\fi\global\qedtrue}

%%
%% The eqnqed/label hack is accomplished by mucking
%% around with some internal LaTeX definitions.  
%%
\makeatletter
\def\eqnqed{\noqed
	\def\@tempa{equation}
	\ifx\@tempa\@currenvir\def\@eqnnum{\qsym}%
	\addtocounter{equation}{-1}\else%
    \def\@@eqncr{\let\@tempa\relax
    \ifcase\@eqcnt \def\@tempa{& & &}\or \def\@tempa{& &}%
      \else \def\@tempa{&}\fi
     \@tempa {\def\@eqnnum{{\qsym}}\@eqnnum}%  [changed theequation to @eqnnum]
     \global\@eqnswtrue\global\@eqcnt\z@\cr}\fi}

%% OLD VERSION
%% 
%% \def\eqnqed{\noqed\@eqnswtrue%
%%     \def\@@eqncr{\let\@tempa\relax
%%     \ifcase\@eqcnt \def\@tempa{& & &}\or \def\@tempa{& &}%
%%       \else \def\@tempa{&}\fi
%%      \@tempa \qsym%
%%      \global\@eqnswtrue\global\@eqcnt\z@\cr}}
%% 

%%
%% The same trick can be used to produce arbitrary labels for
%% equations.
%%
\def\eqnlabel#1#2{\if@filesw {\let\thepage\relax%
   \def\protect{\noexpand\noexpand\noexpand}%
   \edef\@tempa{\write\@auxout{\string
      \newlabel{#2}{{{#1}}{\thepage}}}}%
   \expandafter}\@tempa%
   \if@nobreak \ifvmode\nobreak\fi\fi\fi%
	\def\@tempa{equation}
	\ifx\@tempa\@currenvir\def\theequation{{#1}}% 
	\addtocounter{equation}{-1}\else%
    \def\@@eqncr{\let\@tempa\relax
    \ifcase\@eqcnt \def\@tempa{& & &}\or \def\@tempa{& &}%
      \else \def\@tempa{&}\fi
     \@tempa {\def\@eqnnum{{#1}}\@eqnnum}% [changed theequation to @eqnnum]
     \global\@eqnswtrue\global\@eqcnt\z@\cr}\fi}

\makeatother

%% end of qed/eqnlabel

\newcommand{\listqed}{\qed\noqed} % RJB - use if proof ends with a list
% RJB Use \littleqed at the end of proving assertions inside proofs
\def\littleqed{\ifqed{\penalty1000\unskip\nobreak\hfil\penalty50
\hskip2em\hbox{}\nobreak\hfil\littleqsym
\parfillskip=0pt \finalhyphendemerits=0\par\medskip}\fi\global\qedtrue}
\def\littleqsym{\vrule width0.6ex height0.6em depth0ex}
\def\QED{\qed}

\makeatother

%% end of qed/eqnlabel
%%%%%%%%%%%%%%%%%%%%%%%%%%%%%%%%%%%%%%%%%%%%%%%%%%%%%%%%%%%%%%%%
%%%%%%%%%%%%%%%%%   end of Nick's qed stuff   %%%%%%%%%%%%%%%%%%
%%%%%%%%%%%%%%%%%%%%%%%%%%%%%%%%%%%%%%%%%%%%%%%%%%%%%%%%%%%%%%%%

\def\aftereqnskip{\vskip-3pt} % compositor, please adjust this

%%%%%%%%%%%%%%%%%%%%%%%%%%%%%%%%%%%%%%%%%%%%%%%%%%%%%%%%%%%%%%%%

\newcommand{\SPASAM}[1]{{\rm MOL\dash A}(#1)}
\newcommand{\SPASM}[1]{{\rm MOL'}(#1)}
\newcommand{\SPAM}[1]{{\rm MOL}(#1)}
\newcommand{\NPbits}[1]{{\rm NPbits}(#1)}
\newcommand{\NPinit}[1]{{\rm NPinit}(#1)}
\newcommand{\NPpaths}[1]{{\rm NPpaths}(#1)}
\newcommand{\LOR}{\bigvee}

\newcommand{\set}[1]{\left\{{#1}\right\}}

\title{Arithmetic Simplicity in Stochastic Gradient Methods }

\author{Bin Fu,
 Pengfei Gu, Jose Nunez, and Fabian Vazquez
\\
\\
 Department of Computer Science\\
 University of Texas Rio Grande Valley\\
 Edinburg, TX 78539, USA\\
 %bin.fu@utrgv.edu
 }
%\date{} %April 2024}
\maketitle

\begin{abstract}
A gradient descent method is  arithmetically simple  if the operations are limited to  $+,-, \times$, and  division $x/2^t$ with  integer $t$. An arthmetically simple gradient method is easy to implement in chip design.
We show how to transform AdaGrad, Adam, and AdamW into arithmetically simple. 

AdamW is based on the recursion $x_{t+1}=(1-\lambda\eta)x_t-\frac{\eta }{s}m_t$ and Adam is the special case of AdamW with $\lambda=0$. We transform them into a static case with $s=S(T)$, where $T$ is the number of iterations, and $S(T)$ is a fixed function.   The convergence analysis is given for a static Adam, which is also arithmetically simple.

\end{abstract}

%\end{document}

\section{Introduction}

Stochastic gradient methods have become a fundamental class of optimization algorithms for large-scale machine learning. Given an objective function
$
F(x)$ with a stochastic gradient $G(x,\xi),
$
where $(x\in\mathbb{R}^d)$ denotes the model parameters and $(\xi)$ represents a random sample or mini-batch, stochastic gradient methods iteratively update the parameters using an unbiased or approximately unbiased estimate of the gradient,
$
g_t=\nabla G(x_t,\xi_t),\qquad
x_{t+1}=x_t-\eta_t g_t,
$
where $(\eta_t>0)$ is the learning rate. Compared with deterministic gradient descent, stochastic methods substantially reduce the computational cost of each iteration and are therefore particularly well suited to optimization problems involving massive datasets and high-dimensional parameter spaces.

The classical stochastic approximation framework dates back to Robbins and Monro, while stochastic gradient descent (SGD) subsequently became one of the principal optimization mechanisms for statistical learning. The basic SGD iteration uses a single sample or mini-batch to approximate the full gradient. Although computationally efficient, conventional SGD uses the same learning-rate scale for all coordinates. Consequently, when different coordinates of the objective have substantially different curvatures or gradient magnitudes, selecting an appropriate global learning rate can be difficult.
AdaGrad is based on the iteration  
$x_{t+1}=x_t-\eta
\frac{g_t}{\sqrt{\sum_{s=1}^{t}g_s^2}+\epsilon},
$
where the operations are coordinatewise. Thus, coordinates that have repeatedly received large gradients are assigned smaller effective learning rates, whereas relatively infrequent coordinates can retain larger learning rates. This adaptive treatment of the geometry of the observed data was shown to provide strong theoretical guarantees and significant practical improvements over non-adaptive stochastic subgradient methods~\cite{DuchiHazanSinger2011}.
Adam is based on iteration $v_j = \theta v_{j-1} + (1-\theta)(g_j)^{\odot 2} , m_j = \beta m_{j-1} + (1-\beta)g_j$, $x_{t+1}=x_t-\eta
\frac{ m_t}{\sqrt{ v_t}+\epsilon}.
$
Adam therefore combines momentum-like first-moment estimation with AdaGrad-like adaptive scaling based on second moments. The method is computationally efficient, requires relatively little memory beyond the parameter and moment vectors, and is particularly effective for large-scale problems with sparse or noisy gradients~\cite{KingmaBa2015}.

%Despite its practical success, the treatment of regularization and weight decay in adaptive methods presents an important subtlety. For standard SGD, adding an $(L_2)$ regularization term to the objective can be interpreted equivalently, after an appropriate rescaling, as multiplicative weight decay. For an adaptive method such as Adam, however, this equivalence breaks down because the regularization gradient is itself transformed by the adaptive preconditioning mechanism. In other words, if $(g_t)$ is replaced by $(g_t+\lambda x_t)$, the regularization term is normalized coordinatewise together with the stochastic gradient. Consequently, the resulting operation is not the same as directly shrinking the parameters.
AdamW uses iteration 
$x_{t+1}=(1-\eta\lambda)x_t+\eta
\frac{ m_t}{\sqrt{ v_t}+\epsilon}$ (Adam has $\lambda=0$).
Thus, the adaptive optimization mechanism and the weight-decay mechanism operate independently. Loshchilov and Hutter showed that this decoupling removes an undesirable interaction between the learning rate and the weight-decay coefficient and substantially improves the generalization behavior of Adam on several benchmark problems~\cite{Loshchilov2019Decoupled}.

The progression from SGD to AdaGrad, Adam, and AdamW can therefore be viewed as a sequence of increasingly sophisticated mechanisms for controlling the geometry of stochastic optimization~\cite{DuchiHazanSinger2011,DBLP:journals/corr/abs-1002-4908,NemirovskiJuditskyLanShapiro2009,BottouCurtisNocedal2018,WardWuBottou19,XieWuWard2020,DBLP:journals/corr/abs-1212-5701}. SGD uses a global learning rate; AdaGrad introduces coordinatewise adaptation based on accumulated gradient information; Adam combines adaptive second-moment scaling with momentum through first-moment estimation; and AdamW further separates optimization from parameter regularization. This progression illustrates a central theme in modern stochastic optimization: the performance of a first-order method depends not only on the stochastic gradient itself, but also on how historical gradient information, parameter scaling, and regularization are incorporated into the update rule.

A distinctive feature of the proposed gradient methods in this paper is their arithmetic simplicity,  which is defined in~\cite{Fu2026Adaptivity}.  The algorithms are designed to use only addition, subtraction, multiplication, and restricted division of the form $(x/2^t)$, where $t$ is an integer. This restriction is particularly attractive for hardware implementation because division by a power of two can generally be realized through binary shifting or exponent adjustment, avoiding the complexity of a general-purpose division unit. Consequently, the proposed methods have the potential to map naturally onto FPGA, ASIC, and other specialized accelerator architectures.

Arithmetic simplicity can provide several benefits for chip design. General division and square-root operations typically require more complicated arithmetic circuitry and can introduce additional latency, area, and energy costs compared with simpler arithmetic operations. By eliminating such operations from the optimization procedure, the proposed approach may enable simpler datapaths, reduced hardware complexity, improved computational throughput, and lower energy consumption. %The restricted arithmetic structure may also facilitate fixed-point or reduced-precision implementations, although the effects of finite precision and numerical error must be evaluated carefully.

We show how to transform AdaGrad, Adam, and AdamW into arithmetically simple.
%, and provide simulation results. 
After Adam is transformed into arithmetically simple, it has a faster convergence in our experiments.

AdamW is based on the recursion $x_{t+1}=(1-\lambda\eta)x_t-\frac{\eta }{s}m_t$ and Adam is the special case of AdamW with $\lambda=0$. If $T$ is the number of iterations for a gradient method, we use function $S(T)$ to determine its step size $s=S(T)$. We transform the gradient method into a static case with $\eta=S(T)$, where $T$ is the number of iterations, and $S(T)$ is a fixed function. We have a static Adam with recursion $x_{t+1}=x_t-\frac{\eta }{s}m_t$, and static AdamW with recursion  $x_{t+1}=(1-\lambda\eta)x_t-\frac{\eta }{s}m_t$.

Let $\xi$ be a random variable and $G(\xi, x)$ be a stochastic approximation  for $\bigtriangledown F(x)$.
(1) $\langle \bigtriangledown F(x), \Expect_{\xi}(G(\xi, x))\rangle\ge  \lambda_0 \Vert\bigtriangledown F(x)\Vert^2$ for some $\lambda_0\in (0,1)$, and (2) $\Expect_{\xi}(\Vert \bigtriangledown F(x)-G(\xi, x)\Vert^2)\le\sigma_0^2+\sigma_1^2\Vert \bigtriangledown F(x)\Vert^2$ for some $\sigma_0,\sigma_1\in [0,+\infty)$.  The convergence analysis is given for a   static Adam, which is also arithmetically simple. 
%We show that static Adam has a convergence rate $O\left(\frac{1}{T^{0.5}}\right)$ if $G(\xi, x)$ satisfies (1) $\langle \bigtriangledown F(x), \Expect_{\xi}(G(\xi, x))\rangle\ge  \lambda_0 \Vert\bigtriangledown F(x)\Vert^2$ for some $\lambda_0\in (0,1)$, and (2) $\Expect_{\xi}(\Vert \bigtriangledown F(x)-G(\xi, x)\Vert^2)\le\sigma_0^2+\sigma_1^2\Vert \bigtriangledown F(x)\Vert^2$ for some $\sigma_0,\sigma_1\in [0,+\infty)$.  We also show that static Adam has a convergence rate $O\left(\frac{1}{T^{0.3}}\right)$ if $G(\xi, x)$ satisfies (1) $\Expect_{\xi}(G(\xi, x)=\bigtriangledown F(x)$,   and (2) $\Expect_{\xi}(\Vert \bigtriangledown F(x)-G(\xi, x)\Vert^2)\le\sigma_0^2+\sigma_1^2\Vert \bigtriangledown F(x)\Vert^2$ for some $\sigma_0,\sigma_1\in [0,+\infty)$. %Both static Adam and static AdamW are integrated a parallel framework that was recently developed by Fu. It can automatically adapt  to Lipschitz smoothness constant and stochastic gradient characteristics such as the variance and noise level.

%The momentum gradient descents are integrated into the parallel framework that was recently developed by Fu, and brings a simplified proof about its convergence. We also show parallel simulation about the practical performance.

\section{Preliminaries}

Let $\mathbb{R}=(-\infty,+\infty)$ be the set of real numbers. A vector in $\mathbb{R}^d$ is $(a_1, a_2,\ldots, a_d)$ with $a_i\in \mathbb{R}$ for $i=1,2,\cdots, d$. The dot product between two vectors $V=(v_1,v_2,\ldots, v_d)$ and $U=(u_1,u_2,\ldots, u_d)$ is denoted by $\langle U, V\rangle=\sum_{i=1}^d u_iv_i$. The length of a vector $V=(v_1,v_2,\ldots, v_d)$  is denoted by $\Vert V\Vert=\sqrt{v_1^2+v_2^2+\ldots+v_d^2}$.

\begin{itemize}
    \item 
    {\bf $L$-Lipschitz} smooth: 
    
\begin{eqnarray}
    \Vert \bigtriangledown(F(x))-\bigtriangledown(F(y))\Vert\le L\Vert x-y\Vert.\label{Lipsc-ineqn}    
\end{eqnarray}

%\item
%{\bf $\mu$-Polyak-Lojasjewicz} Inequality: $\Vert \bigtriangledown(F(x))\Vert^2\ge 2\mu(F(x)-F(x^*))$. 

\item $F^*=\inf_{x} F(x)>-\infty$.

\end{itemize}
Let $C_L^1$ be the class of functions that are $L$-Lipschitz smooth.
The following Lemma~\ref{basic-lemma} can be easily proven by $L$-Lipschitz condition and Taylor expansion. 

\begin{lemma}\label{basic-lemma}
    Let $F(x_1,\cdots, x_d)$ be a function $\mathbb{R}^d\rightarrow \mathbb{R}$ in $C_L^1$, we have $F(x)\le F(y)+\langle\bigtriangledown F(y), x-y\rangle+{L\over 2}\Vert x-y\Vert^2$.
\end{lemma}

\begin{definition}\label{condition} Let $\xi$ be a random variable and $G(\xi, x)$ be an approximation for $\bigtriangledown F(x)$
	\begin{enumerate}
		\item\label{gradient-variance-condition} $\langle\bigtriangledown F(x),\Expect_{\xi}(G(\xi, x))\rangle\ge  \lambda_0 \Vert\bigtriangledown F(x)\Vert^2$ for some $\lambda_0\in (0,1)$.
		\item\label{condition2-def} 
		$\Expect_{\xi}(\Vert \bigtriangledown F(x)-G(\xi, x)\Vert^2)\le\sigma_0^2+\sigma_1^2\Vert \bigtriangledown F(x)\Vert^2$ for some $\sigma_0,\sigma_1\in [0,+\infty)$.
	\end{enumerate}
	
\end{definition}

\begin{definition}
  A gradient descent method is {\it arithmetically simple} if the operations are limited to  $+,-, \times$, and  division $x/2^t$ with  integer $t$.
\end{definition}

\section{Algorithms }

In this section we give a description for a new adaptive gradient descent algorithm.

%\vskip 20pt
%{\bf Algorithm} Static-Momentum-GD$(G(.,.), \alpha,\beta, \eta,  x_0, t, T)$

%Input: 

%\begin{itemize}
%	\item
%	$G(\xi, x):\mathbb{R}^m\rightarrow \mathbb{R}$ is an approximation for $\bigtriangledown F(x)$,

	%    \item $\epsilon\in (0,1)$,
	
	%    \item $b_0\in [1,+\infty)$, 

%\item$\alpha \in (0,+\infty)$ determines the weight of the current gradient,

%\item $\beta\in [0,1)$ determine the weight of history momentum,

%	\item $\eta\in (0,+\infty)$ control the initial step rate, 
    
%	\item $x_0\in \mathbb{R}^m$ is the start point, 
	
%	\item $t$ is an integer to control rate, 
	
%	\item $T$ is for the number of steps 
%\end{itemize}

%Steps:

%\begin{enumerate}
	
	% Let $s_0=2^h$ with $h=\ceiling{1 \over 2\epsilon}$.
	
%	\item Let $x_{1}=x_0$
	
%	\item Let $m_{0}=0_m$ (zero vector in $\mathbb{R}^m)$
	
%	\item Let $s_t=2^t$
	
%	\item Let $j=1$

%	\item Repeat
	
%	\item $\{$ 

%	\item \qquad Let $g_j=G(\xi_{j},x_{j})$
	
	%\item\qquad\label{moment-line}
%	 Let $m_{j}=\beta  m_{j-1}+\alpha g_j$

%	\item\label{x-j-line} \qquad
%	Let  $x_{j+1}=x_{j}-{\eta\over s_t}\cdot m_{j}$ 

%	\item \qquad Let $j=j+1$
	
%	\item $\}$
	
%	\item Until $j=T$
	%\end{enumerate}

%\end{enumerate}

%{\bf End of Algorithm}

Input parameters for the AdamW:

\begin{itemize}
	\item
	$G(\xi, x):\mathbb{R}^m\rightarrow \mathbb{R}$ is an approximation for $\bigtriangledown F(x)$,

%\item$\alpha \in (0,+\infty)$ determines the weight of the current gradient,

\item $\beta\in [0,1)$ determine the weight of history momentum,

	\item $\theta\in [0,1)$
    
	\item $\eta\in (0,+\infty)$ controls the initial step rate, 

    \item $\lambda\in (0,1)$ is weight-decay coefficient,
    
	\item $x_0\in \mathbb{R}^m$ is the start point, 
	
%	\item $t$ is an integer to control rate, 
	
	\item $T$ is for the number of steps 
\end{itemize}

\noindent
\begin{table}[htbp]
\centering
\caption{Side-by-side comparison of AdamW and Generalized-AdamW.}
\label{tab:alg_merge}
\begin{tabularx}{\textwidth}{l X X}
\toprule
\textbf{Step} & \textbf{Algorithm AdamW} & \textbf{Algorithm Generalized-AdamW} \\
\midrule
\textbf{Input} & 
$G(\xi,x)$, $\beta, \theta, \eta, \lambda, x_0, T$ & 
$G(\xi,x)$, $S(\cdot), \beta, \theta, \eta, \lambda, x_0, T$ \\
\midrule
1. & Let $x_1 = x_0$ & Let $x_1 = x_0$ \\
2. & Let $m_0 = 0_m$ (zero vector in $\mathbb{R}^m$) & Let $m_0 = 0_m$ (zero vector in $\mathbb{R}^m$) \\
3. & Let $v_0 = 0$ & Let $v_0 = 0$ \\
4. & Let $j = 1$ & Let $j = 1$ \\
5. & \textbf{Repeat} & \textbf{Repeat} \\
6. & \quad Let $g_j = G(\xi_j, x_j)$ & \quad Let $g_j = G(\xi_j, x_j)$ \\
7. & \quad Let $m_j = \beta m_{j-1} + (1-\beta)g_j$ & \quad Let $m_j = \beta m_{j-1} + (1-\beta)g_j$ \\
8. & \quad Let $v_j = \theta v_{j-1} + (1-\theta)(g_j)^{\odot 2}$ & \quad Let $v_j = \theta v_{j-1} + (1-\theta)(g_j)^{\odot 2}$ \\
9. & \quad \textit{(Standard denominator $\sqrt{v_j+\epsilon}$)} & \quad Let $s = S(v_j, T)$ \\
10. & \quad $x_{j+1} = (1-\lambda\eta)x_j - \frac{\eta}{\sqrt{v_j+\epsilon}} \odot m_j$ & \quad $x_{j+1} = (1-\lambda\eta)x_j - \frac{\eta}{s} \odot m_j$ \\
11. & \quad Let $j = j+1$ & \quad Let $j = j+1$ \\
12. & \textbf{Until} $j = T$ & \textbf{Until} $j = T$ \\
\bottomrule.
\end{tabularx}
\end{table}

\begin{definition}In generalized-AdamW, we define the special cases:
\begin{itemize}
    \item Static-Adam:  $\lambda=0$ and $S(v_j, T)$ only depends on $T$.
        \item Static-AdamW:  $S(v_j, T)$ only depends on $T$.

    \item AdaGrad: $\lambda=0, \beta=0$, and $S(v_j,T)=\sqrt{v_j+\epsilon}$.

     \item Adam: $\lambda=0$, and $S(v_j,T)=\sqrt{v_j+\epsilon}$.

         \item AdamW: $S(v_j,T)=\sqrt{v_j+\epsilon}$.
    
    \item Arithmetically-Simple-AdaGrad:  $\lambda=0, \beta=0$, and $S(v_j,T)=2^t$ with integer $t$ and $2^t\in [\frac{1}{2}\sqrt{v_j+\epsilon}, 2\sqrt{v_j+\epsilon}]$.
    \item Arithmetically-Simple-Adam:  $\lambda=0$ and $S(v_j,T)=2^t$ with integer $t$ and $2^t\in [\frac{1}{2}\sqrt{v_j+\epsilon}, 2\sqrt{v_j+\epsilon}]$.
        \item Arithmetically-Simple-AdamW:   $S(v_j,T)=2^t$ with integer $t$ and $2^t\in [\frac{1}{2}\sqrt{v_j+\epsilon}, 2\sqrt{v_j+\epsilon}]$.
\end{itemize}

\end{definition}

To bridge standard adaptive algorithms and hardware-friendly implementations, AdamW and its variants can be transformed into \textbf{Arithmetically Simple} counterparts. Specifically, let $x \in \mathbb{R}^+$ be a positive floating-point number. We define a transformation function $S(x)$ that maps $x$ to a power of two, $2^t$ for some integer $t$, such that $2^t \in [\frac{1}{2}\sqrt{x}, 2\sqrt{x}]$. 

For instance, following the IEEE standard for floating-point representation, a positive real number can be expressed as $x = a \times 2^m$ with $a \in [1, 2)$. Setting $S_C(x) = 2^{\lceil(m+1)/2\rceil}$ ensures rigorous enclosure bounds: since $x \le 2 \times 2^m = 2^{m+1}$, we obtain:
$$\sqrt{x} \le 2^{(m+1)/2} \le S_C(x)$$
and 
$$S_C(x) = 2^{\lceil(m+1)/2\rceil} \le 2^{(m+2)/2} = 2 \times 2^{m/2} \le 2\sqrt{x}.$$

Similiaryly, if we let $S_F(x)=2^{\lfloor(m+1)/2\rfloor}$, we have $\frac{1}{2}\sqrt{x}\le S_F(x)\le \sqrt{x}$.

The \textbf{Arithmetically Simple AdamW} algorithm modifies standard AdamW by replacing the square-root denominator term $\sqrt{v_j + \epsilon}$ elementwise with $S(y_j)$, where $y_j$ represents the $j$-th element of the second-moment vector $v_j$. The arithmetically simple versions of AdaGrad and Adam are derived analogously. Therefore, we have

\begin{proposition}
 In generalized-AdamW, let $S(x)=S_C(x)$ or $S(x)=S_F(x)$, then we have
\begin{itemize}       
    \item If  $\lambda=0,$ and $ \beta=0$, then it is Arithmetically-Simple-AdaGrad.
    \item If  $\lambda=0$, then it is Arithmetically-Simple-Adam. 
        \item It is Arithmetically-Simple-AdamW.  
\end{itemize}  
\end{proposition}

\section{Some Technical Lemmas}

In this section, we show some technical lemmas. They will be used in convergence analysis for the following gradient method that is arithmetically simple. It can be transformed into a Static-Adam by selecting a  $S(.)$.

\vskip 20pt
{\bf Algorithm} Static-Momentum-GD$(G(.,.), \alpha,\beta, \eta,  x_0, t, T)$

Input: 

\begin{itemize}
	\item
	$G(\xi, x):\mathbb{R}^m\rightarrow \mathbb{R}$ is an approximation for $\bigtriangledown F(x)$,

	%    \item $\epsilon\in (0,1)$,
	
	%    \item $b_0\in [1,+\infty)$, 

\item$\alpha \in (0,+\infty)$ determines the weight of the current gradient,

\item $\beta\in [0,1)$ determine the weight of history momentum,

	\item $\eta\in (0,+\infty)$ control the initial step rate, 
    
	\item $x_0\in \mathbb{R}^m$ is the start point, 
	
	\item $t$ is an integer to control rate, 
	
	\item $T$ is for the number of steps 
\end{itemize}

Steps:

\begin{enumerate}
	
	% Let $s_0=2^h$ with $h=\ceiling{1 \over 2\epsilon}$.
	
	\item Let $x_{1}=x_0$
	
	\item Let $m_{0}=0_m$ (zero vector in $\mathbb{R}^m)$
	
	\item Let $s_t=2^t$
	
	\item Let $j=1$

	\item Repeat
	
	\item $\{$

	\item \qquad Let $g_j=G(\xi_{j},x_{j})$
	
	\item\qquad\label{moment-line}
	 Let $m_{j}=\beta  m_{j-1}+\alpha g_j$

	\item\label{x-j-line} \qquad
	Let  $x_{j+1}=x_{j}-{\eta\over s_t}\cdot m_{j}$

	\item \qquad Let $j=j+1$
	
	\item $\}$
	
	\item Until $j=T$
	%\end{enumerate}

\end{enumerate}

{\bf End of Algorithm}

\begin{lemma}
	\label{recur-lemma}	
	Assume that $m_j$ and $x_j$ are generated by the algorithm. Then we have
	\begin{enumerate}
	\item\label{m-j-case} 
		$m_1=0_m$, and $m_{j}=\sum_{i=1}^j\alpha\beta^{j-i}g_i$ for all $j\ge 0$.		
		\item $x_1=x_0$, and 
	$x_{j+1}-x_j=-{\eta\over s_t}m_{j}$		for all $j\ge 0$.
	\end{enumerate}

\end{lemma}

\begin{proof} We give an inductive proof. By the algorithm, we have $m_2=\alpha g_1=\sum_{i=1}^1\alpha\beta^{1-1}g_i$. This proves the case $j=1$. Assume that $m_{j}=\sum_{i=1}^j\alpha\beta^{j-i}g_i$. By the algorithm, we have 
\begin{eqnarray*}
    m_{(j+1)}&=&\beta m_{j}+\alpha g_{j+1}\\
    &=&\beta (\sum_{i=1}^j\alpha\beta^{j-i}g_i)+\alpha g_{j+1}\\
    &=& (\sum_{i=1}^j\alpha\beta^{(j+1)-i}g_i)+\alpha g_{j+1}\\
    &=& (\sum_{i=1}^{j+1}\alpha\beta^{(j+1)-i}g_i).
\end{eqnarray*}
This proves (\ref{m-j-case}) of this lemma. For the second case of this lemma, 
it follows from line~(\ref{x-j-line}) of the algorithm.

\end{proof}

\begin{lemma}\label{basic-sum-lemma}
   Let $a_0,a_1,\ldots$ be a series of nonnegative real numbers, and $b\in (0,1]$. Then (1)  $\sum_{i=1}^j\sum_{t=1}^j b^{j-i-t}a_t\le \sum_{i=1}^j  \frac{b^{-t}}{1-b}a_t$,  (2) $\sum_{i=1}^j\sum_{t=1}^i b^{i-t}a_t\le \sum_{t=1}^j  \frac{1}{1-b}a_t$, (3) $\sum_{j=1}^T\sum_{i<j, j-i>v} b^{j-i-v}a_t\le \sum_{i=1}^T  \frac{b}{1-b}a_i$.
\end{lemma}

\begin{proof}
(1) We have the inequalities:
   \begin{eqnarray*}
   &&\sum_{i=1}^j\sum_{t=1}^j b^{j-i-t}a_t=\sum_{t=1}^j\sum_{i=1}^j b^{j-i-t}a_t\\
   &=&\sum_{t=1}^ja_tb^{-t}\sum_{i=1}^j b^{j-i} 
   \le \sum_{t=1}^j  \frac{b^{-t}}{1-b}a_t
   \end{eqnarray*}
   
(2) For the second case, we have inequalities:
 \begin{eqnarray*}
     &&\sum_{i=1}^j\sum_{t=1}^i b^{i-t}a_t
\le\sum_{t=1}^j\sum_{i=t}^j b^{i-t}a_t\\
     &\le&\sum_{t=1}^ja_t\sum_{i=t}^j b^{i-t}\le \sum_{t=1}^j  \frac{1}{1-b}a_t.
 \end{eqnarray*}

 %(3) 
% \begin{eqnarray*}
%&&\sum_{j=1}^T\sum_{i<j, j-i>v} b^{j-i-v}a_i=\sum_{j=1}^T\sum_{i=j-v-1}^{j-1} b^{j-i-v}a_i\\
%&\le&\sum_{j=1}^T\sum_{i=j-v-1}^{T} b^{j-i-v}a_i\le\sum_{i=1}^Ta_i\sum_{j=1}^{T} b^j\le \sum_{i=1}^T  \frac{b}{1-b}a_i.    
% \end{eqnarray*}
%We note that $1\le j-i-v\le T$.

\end{proof}

%$\frac{a}{(1-a)^2}$
%by Lemma~\ref{basic2-sum-lemma}

\begin{lemma}\label{basic2-sum-lemma}
  For a real $a\in [0,1)$, $\sum_{i=1}^{+\infty}ia^i=\frac{a}{(1-a)^2}$.  
\end{lemma}

\begin{proof}
  Let $f(x)=\sum_{i=0}^{+\infty}x^i=\frac{1}{1-x}$. We have derivative  $f(x)'=\frac{1}{(1-x)^2}=\sum_{i=1}^{+\infty}i\cdot x^{i-1}$. Therefore, $\sum_{i=1}^{+\infty}i\cdot x^{i}=xf(x)'=\frac{x}{(1-x)^2}$.
\end{proof}

\begin{lemma}\label{basic0-lemma}
$\Vert x_{j+1}-x_j\Vert^2	\le {1\over 1-\beta}\left({\alpha\eta\over s_t}\right)^2(\sum_{i=1}^j\beta^{j-i}(\Vert g_i\Vert^2)$.
\end{lemma}

\begin{proof}By the algorithm, we have $x_{j+1}-x_j=-{\eta\over s_t}m_{j+1}$.
	
	We have the inequalities: 
	\begin{eqnarray*}
		&&\Vert x_{j+1}-x_j\Vert^2=\langle  -{\eta\over s_t}m_{j}, -{\eta\over s_t}m_{j}\rangle=\left({\eta\over s_t}\right)^2\langle m_{j}, m_{j}\rangle\\	
&=&\left({\eta\over s_t}\right)^2\langle \sum_{i=1}^j\alpha\beta^{j-i}g_i, \sum_{i=1}^j\alpha\beta^{j-i}g_i\rangle\ \ (by\ Lemma~\ref{recur-lemma})\\
&=&\left({\eta\over s_t}\right)^2\sum_{i=1}^j\sum_{t=1}^j\langle \alpha\beta^{j-i}g_i, \alpha\beta^{j-i}g_t\rangle=\left({\eta\over s_t}\right)^2\sum_{i=1}^j\sum_{t=1}^j \alpha^2\beta^{j-i+j-t}\langle g_i, g_t\rangle\\
&=&\left({\alpha\eta\over s_t}\right)^2\sum_{i=1}^j\sum_{t=1}^j \beta^{2j-i-t}\langle g_i, g_t\rangle\\\
&\le&\left({\alpha\eta\over s_t}\right)^2\sum_{i=1}^j\sum_{t=1}^j \beta^{2j-i-t}(\Vert g_i\Vert \cdot \Vert g_t\Vert)\ \ (by\ Cauchy-Schwarz\  inequality)\\
\\
&\le&{1\over 2}\left({\alpha\eta\over s_t}\right)^2\sum_{i=1}^j\sum_{t=1}^j \beta^{2j-i-t}(\Vert g_i\Vert^2+ \Vert g_t\Vert^2)\\
&\le&{1\over 2}\left({\alpha\eta\over s_t}\right)^2\left(\left(\sum_{i=1}^j\sum_{t=1}^j \beta^{2j-i-t}\Vert g_i\Vert^2\right)+ \left(\sum_{i=1}^j\sum_{t=1}^j \beta^{2j-i-t}\Vert g_t\Vert^2\right)\right)\\
&\le&{1\over 2}\left({\alpha\eta\over s_t}\right)^2\left(\left(\sum_{i=1}^j\beta^{j-i}\Vert g_i\Vert^2\right)\sum_{t=1}^j \beta^{j-t}+ \left(\sum_{i=1}^j\sum_{t=1}^j \beta^{2j-i-t}\Vert g_t\Vert^2\right)\right)\\
&\le&{1\over 2}\left({\alpha\eta\over s_t}\right)^2\left(\left(\sum_{i=1}^j\beta^{j-i}\Vert g_i\Vert^2\right){1\over 1-\beta}+ \left(\sum_{i=1}^j\sum_{t=1}^j \beta^{2j-i-t}\Vert g_t\Vert^2\right)\right)\\
&\le&{1\over 2}\left({\alpha\eta\over s_t}\right)^2\left(\left(\sum_{i=1}^j\beta^{j-i}\Vert g_i\Vert^2\right){1\over 1-\beta}+ \beta^{j}\left(\sum_{i=1}^j\sum_{t=1}^j \beta^{j-i-t}\Vert g_t\Vert^2\right)\right)\\
&\le&{1\over 2}\left({\alpha\eta\over s_t}\right)^2\left(\left(\sum_{i=1}^j\beta^{j-i}\Vert g_i\Vert^2\right){1\over 1-\beta}+ \left(\sum_{i=1}^j \beta^{j-i}\Vert g_i\Vert^2\right){1\over 1-\beta}\right)\\
&&(by\ (1)\ of\  Lemma~\ref{basic-sum-lemma})\\
&=&{1\over 1-\beta}\left({\alpha\eta\over s_t}\right)^2\left(\sum_{i=1}^j\beta^{j-i}\Vert g_i\Vert^2\right).
\end{eqnarray*}
\end{proof}

%***

\begin{lemma}\label{basic0b1-lemma}Assume 
\begin{eqnarray*}
&&\Vert x_{j+1}-x_j\Vert^2	\le C\sum_{i=1}^j\beta^{j-i}\Vert g_i\Vert^2.    
\end{eqnarray*}
Then we have 
\begin{eqnarray*}
    \sum_{z=u}^v\Vert x_{z+1}-x_z\Vert^2\le \frac{C}{1-\beta}\cdot \sum_{i=1}^v\Vert g_i\Vert^2.
\end{eqnarray*}
\end{lemma}

\begin{proof}
    \begin{eqnarray*}
    \sum_{z=u}^v\Vert x_{z+1}-x_z\Vert^2&\le& \sum_{z=u}^v\left(C\cdot \sum_{i=1}^z\beta^{z-i}\Vert g_i\Vert^2\right)\\
&\le&  C\cdot \sum_{i=1}^v\sum_{z=1}^z\beta^{z-i}\Vert g_i\Vert^2\\
&=&  C\cdot \sum_{i=1}^v\Vert g_i\Vert^2\sum_{z=1}^z\beta^{z-i}\\
&\le&  C\cdot \sum_{i=1}^v\Vert g_i\Vert^2\cdot \frac{1}{1-\beta}\\
&=&  \frac{C}{1-\beta}\cdot \sum_{i=1}^v\Vert g_i\Vert^2.
\end{eqnarray*}
\end{proof}

\section{Convergence in Stochastic Model}

%\end{document}

%\begin{definition}\label{condition} Let $\xi$ be a random variable and $G(\xi, x)$ be an approximation for $\bigtriangledown F(x)$
%	\begin{enumerate}
		%\item\label{gradient-variance-condition} $\langle\bigtriangledown F(x),\Expect_{\xi}(G(\xi, x))\rangle\ge  \lambda_0 \Vert\bigtriangledown F(x)\Vert^2$ for some $\lambda_0\in (0,1)$.
		%\item\label{condition2-def} 
%		$\Expect_{\xi}(\Vert \bigtriangledown F(x)-G(\xi, x)\Vert^2)\le\sigma_0^2+\sigma_1^2\Vert \bigtriangledown F(x)\Vert^2$ for some $\sigma_0,\sigma_1\in [0,+\infty)$.
%	\end{enumerate}
	
%\end{definition}

%\end{document}

\begin{lemma}\label{First-Basic-lemma}
	Assume $F(x)$ and $G(\xi, x)$ satisfy %the condition~(\ref{condition2-def})  in Definition~\ref{condition}. 
	$\Expect_{\xi}(\Vert \bigtriangledown F(x)-G(\xi, x)\Vert^2)\le\sigma_0^2+\sigma_1^2\Vert \bigtriangledown F(x)\Vert^2$ for some $\sigma_0,\sigma_1\in [0,+\infty)$.
	Then $\Expect_{\xi}(\Vert G(\xi, x)\Vert^2)\le 2\sigma_0^2+ (2+2\sigma_1^2)\Vert \bigtriangledown F(x)\Vert^2$.
\end{lemma}

\begin{proof} By inequality $(a+b)^2\le 2(a^2+b^2)$, we have
	\begin{eqnarray}
		\Vert G(\xi, x)\Vert^2\le 2\Vert G(\xi, x)-\bigtriangledown F(x)\Vert^2+2\Vert \bigtriangledown F(x)\Vert^2.
	\end{eqnarray}
	Therefore,
	
	\begin{eqnarray}
		\Expect_{\xi}(\Vert G(\xi, x)\Vert^2)&\le&\Expect( 2\Vert G(\xi, x)-\bigtriangledown F(x)\Vert^2+2\Vert \bigtriangledown F(x)\Vert^2)\\
		&=&2\Expect( \Vert G(\xi, x)-\bigtriangledown F(x)\Vert^2)+2\Vert \bigtriangledown F(x)\Vert^2\\
		&\le&2(\sigma_0^2+\sigma_1^2\Vert\bigtriangledown F(x)\Vert^2)+2\Vert \bigtriangledown F(x)\Vert^2\\
		&=&2\sigma_0^2+(2+2\sigma_1^2)\Vert\bigtriangledown F(x)\Vert^2.
	\end{eqnarray}
	
\end{proof}

\begin{lemma}\label{basic2b1-lemma} Assume that $u$ and $v$. For $i<j$,  we have 
\begin{eqnarray*}
\langle \bigtriangledown F(x_j), g_i\rangle&\ge& \langle\bigtriangledown F(x_i), g_i\rangle-(j-i)\Vert g_i\Vert^2-{L\over 2}{1\over (1-\beta)^2}\left({\alpha\eta\over s_t}\right)^2\sum_{p=1}^{j-1}\Vert g_p\Vert^2.	    
\end{eqnarray*}

\end{lemma}

\begin{proof}
We have

\begin{eqnarray*}
   \langle \bigtriangledown F(x_j), g_i\rangle
   &=& \langle (\bigtriangledown F(x_j)-\bigtriangledown F(x_{j-1}))+(\bigtriangledown F(x_{j-1})-\bigtriangledown F(x_{j-2}))+\ldots\\
   &&+(\bigtriangledown F(x_{i+1})-\bigtriangledown F(x_{i}))+\bigtriangledown F(x_i), g_i\rangle\\
   &=& \langle \bigtriangledown F(x_j)-\bigtriangledown F(x_{j-1}), g_i\rangle+\langle\bigtriangledown F(x_{j-1})-\bigtriangledown F(x_{j-2}), g_i\rangle+\ldots\\
   &&+\langle\bigtriangledown F(x_{i+1})-\bigtriangledown F(x_{i}),g_i\rangle\\
&&+\langle\bigtriangledown F(x_i), g_i\rangle\\
   &\ge& -\Vert\langle \bigtriangledown F(x_j)-\bigtriangledown F(x_{j-1})\Vert\cdot \Vert g_i\Vert-\Vert\bigtriangledown F(x_{j-1})-\bigtriangledown F(x_{j-2})\Vert\cdot\Vert g_i\Vert-\ldots\\
   &&-\Vert\bigtriangledown F(x_{i+1})-\bigtriangledown F(x_{i})\Vert\cdot \Vert g_i\Vert+\langle\bigtriangledown F(x_i), g_i\rangle\\
&&(by\ Cauchy\ Schwarz\ inequality)\\
   &\ge& -L\Vert x_j-x_{j-1}\Vert\cdot \Vert g_i\Vert-L\Vert x_{j-1}-x_{j-2}\Vert\cdot\Vert g_i\Vert+\ldots\\
   &&-L\Vert x_{i+1}-x_{i}\Vert\cdot \Vert g_i\Vert+\langle\bigtriangledown F(x_i), g_i\rangle\\
&&(by\ inequality\ (\ref{Lipsc-ineqn}))\\
   &\ge& -{L\over 2}(\Vert x_j-x_{j-1}\Vert^2+ \Vert g_i\Vert^2+\Vert x_{j-1}-x_{j-2}\Vert^2+\Vert g_i\Vert^2+\ldots\\
   &&+\Vert x_{i+1}-x_{i} \Vert^2+ \Vert g_i\Vert^2)+\langle\bigtriangledown F(x_i), g_i\rangle\\
   &=& -{L\over 2}(\Vert x_j-x_{j-1}\Vert^2+\Vert x_{j-1}-x_{j-2}\Vert^2+\ldots+\Vert x_{i+1}-x_{i}\Vert^2)\\
   &&- (j-i)\Vert g_i\Vert^2+\langle\bigtriangledown F(x_i), g_i\rangle\\
   &\ge& -\left({L\over 2}\sum_{t=i}^{j-1}\Vert (x_{t+1}-x_{t})\Vert^2\right)-(j-i)\Vert g_i\Vert^2+\langle\bigtriangledown F(x_i), g_i\rangle\\
      &\ge& \langle\bigtriangledown F(x_i), g_i\rangle-(j-i)\Vert g_i\Vert^2-{L\over 2}{1\over (1-\beta)^2}\left({\alpha\eta\over s_t}\right)^2\sum_{p=1}^{j-1}\Vert g_p\Vert^2\\ 
   &&(By~ Lemma~\ref{basic0-lemma}~and~ Lemma~\ref{basic0b1-lemma}).
\end{eqnarray*}

\end{proof}

\begin{lemma}\label{stochastic2-foundation-lemma2a}For each $1\le j\le T$, we have
\begin{eqnarray*}
 &&\Expect(\langle \bigtriangledown F(x_j), x_{j+1}-x_j\rangle))\le -{\eta\over s_t}(\alpha \Expect(\Vert \bigtriangledown F(x_j)\Vert^2))\\
&&+{\eta\alpha\over s_t}\sum_{i<j}(j-i)\beta^{j-i} \left( \Expect(\Vert g_i\Vert^2)\right)+{\alpha L\over 2(1-\beta)^2}\left({\alpha\eta\over s_t}\right)^3{\beta\over 1-\beta}\left(\sum_{p=1}^{j-1}\Expect(\Vert g_p\Vert^2)\right).
\end{eqnarray*}

\end{lemma}

\begin{proof}
	By Lemma~\ref{recur-lemma}, we have  $x_{j+1}=x_0-{\eta\over s_t}\sum_{i=1}^j\beta(1-\beta)^{j-i}g_i,$
		and $x_{j+1}-x_j=-{\eta\over s_t}m_{j+1}$.
	We have inequalities
		\begin{eqnarray*}
	&&\langle \bigtriangledown F(x_j), x_{j+1}-x_j\rangle\\
    &=&\langle \bigtriangledown F(x_j), -{\eta\over s_t}m_{j}\rangle\\
	&=&-{\eta\over s_t}\langle \bigtriangledown F(x_j), m_{j}\rangle\\
	&=&-{\eta\over s_t}\langle \bigtriangledown F(x_j),\sum_{i=1}^j\alpha\beta^{j-i}g_i\rangle\ \ \ (by~ Lemma~\ref{recur-lemma})	\\
	&=&-{\eta\over s_t}\sum_{i=1}^j\alpha\beta^{j-i} \langle \bigtriangledown F(x_j),g_i\rangle \\		
		&=&-{\eta\over s_t}(\alpha\langle \bigtriangledown F(x_j),g_j\rangle)-{\eta\over s_t}\sum_{i=1}^{j-1}\alpha\beta^{j-i} \langle \bigtriangledown F(x_j),g_i\rangle.
%        &&-{\eta\over s_t}\sum_{i<j, j-i\le v}\alpha\beta^{j-i} \langle \bigtriangledown F(x_j),g_i\rangle\\
\end{eqnarray*}

Therefore, 
\begin{eqnarray*}
	&&\Expect(\langle \bigtriangledown F(x_j), x_{j+1}-x_j\rangle)\\
		&\le&-{\eta\over s_t}(\alpha\lambda_0 \Expect(\Vert \bigtriangledown F(x_j)\Vert^2))-{\eta\over s_t}\sum_{i=1}^{j-1}\alpha\beta^{j-i} \Expect(\langle \bigtriangledown F(x_j),g_i\rangle)\\
        &&(by\ Condition\  (\ref{gradient-variance-condition})~ in~ Definition~\ref{condition} )\\
		&\le&-{\eta\over s_t}(\alpha \Expect(\Vert \bigtriangledown F(x_j)\Vert^2))\\
&&-{\eta\over s_t}\sum_{i=1}^{j-1}\alpha\beta^{j-i} \left( \langle\bigtriangledown F(x_i), \Expect(g_i)\rangle-(j-i)\Expect(\Vert g_i\Vert^2)-{L\over 2}{1\over (1-\beta)^2}\left({\alpha\eta\over s_t}\right)^2\sum_{p=1}^{j-1}\Expect(\Vert g_p\Vert^2)\right)\\
&&(by\ Lemma~\ref{basic2b1-lemma}
%\ and\ Lemma~\ref{basic2b2-lemma}
)\\
        &\le&-{\eta\over s_t}(\alpha\lambda_0 \Expect(\Vert \bigtriangledown F(x_j)\Vert^2))\\
&&+{\eta\over s_t}\sum_{i=1}^{j-1}\alpha\beta^{j-i} \left(-\lambda_0\Vert\bigtriangledown F(x_i)\Vert^2+ (j-i)\Expect(\Vert g_i\Vert^2)+{L\over 2}{1\over (1-\beta)^2}\left({\alpha\eta\over s_t}\right)^2\sum_{p=1}^{j-1}\Expect(\Vert g_p\Vert^2)\right)\\
		&\le&-{\eta\over s_t}(\alpha\lambda_0 \Expect(\Vert \bigtriangledown F(x_j)\Vert^2))\\
&&+{\eta\over s_t}\sum_{i=1}^{j-1}\alpha\beta^{j-i} \left( (j-i)\Expect(\Vert g_i\Vert^2)\right)+{\eta\over s_t}\sum_{i<j}\alpha\beta^{j-i}\left({L\over 2}{1\over (1-\beta)^2}\left({\alpha\eta\over s_t}\right)^2\sum_{p=1}^{j-1}\Expect(\Vert g_p\Vert^2)\right)\\
		&=&-{\eta\over s_t}(\alpha\lambda_0 \Expect(\Vert \bigtriangledown F(x_j)\Vert^2))\\
&&+{\eta\alpha\over s_t}\sum_{i=1}^{j-1}(j-i)\beta^{j-i} \left( \Expect(\Vert g_i\Vert^2)\right)+{\alpha L\over 2(1-\beta)^2}\left({\alpha\eta\over s_t}\right)^3\sum_{i<j}\beta^{j-i}\left(\sum_{p=1}^{j-1}\Expect(\Vert g_p\Vert^2)\right)\\
		&\le&-{\eta\over s_t}(\alpha\lambda_0 \Expect(\Vert \bigtriangledown F(x_j)\Vert^2))\\
&&+{\eta\alpha\over s_t}\sum_{i=1}^{j-1}(j-i)\beta^{j-i} \left( \Expect(\Vert g_i\Vert^2)\right)+{\alpha L\over 2(1-\beta)^2}\left({\alpha\eta\over s_t}\right)^3{\beta\over 1-\beta}\left(\sum_{p=1}^{j-1}\Expect(\Vert g_p\Vert^2)\right).
\end{eqnarray*}
\end{proof}

%***
We need the following inequalities. Let $\gamma_0$ be a  constant in $(0,\frac{1}{2})$.

\begin{eqnarray}   
&&(2+2\sigma_1^2)\left({\eta\beta\alpha \over ( 1-\beta)^2} \right) <\gamma_0{\eta\alpha\lambda_0 }\label{convergence1a-ineqn}\\
&&(2+2\sigma_1^2)\left({\eta\beta\alpha \over s_t( 1-\beta)^2} +{\alpha\beta L\over 2(1-\beta)^3}\left({\alpha\eta\over s_t}\right)^3T\right) \le {\gamma_0\eta\alpha\lambda_0 \over s_t}\label{convergence1a-ineqn2}
\end{eqnarray}

\begin{lemma}\label{stochastic-foundation-lemma2a} Assume that inequality (\ref{convergence1a-ineqn}), we have
\begin{eqnarray*}
 &&\sum_{j=1}^T\Expect(\langle \bigtriangledown F(x_j), x_{j+1}-x_j\rangle))\\
 &\le& -{(1-\gamma_0)\eta\alpha\lambda_0 \over s_t}\sum_{j=1}^T\Expect(\Vert \bigtriangledown F(x_j)\Vert^2)+2\sigma_0^2\left({\eta\beta\alpha \over s_t( 1-\beta)^2} +{\alpha\beta L\over 2(1-\beta)^3}\left({\alpha\eta\over s_t}\right)^3T\right).   
\end{eqnarray*}

\end{lemma}

\begin{proof}
We have
\begin{eqnarray*}
 &&\sum_{j=1}^T\Expect(\langle \bigtriangledown F(x_j), x_{j+1}-x_j\rangle))\\
 &\le& -{\eta\over s_t}(\alpha\lambda_0 \sum_{j=1}^T\Expect(\Vert \bigtriangledown F(x_j)\Vert^2)\\
&&+{\eta\alpha\over s_t}\sum_{j=1}^T\sum_{i=1}^{j-1}(j-i)\beta^{j-i} \left( \Expect(\Vert g_i\Vert^2)\right)+{\alpha L\over 2(1-\beta)^2}\left({\alpha\eta\over s_t}\right)^3{\beta\over 1-\beta}\sum_{j=1}^T\left(\sum_{p=1}^{j-1}\Expect(\Vert g_p\Vert^2)\right)\\
&&(By\ Lemma~\ref{stochastic2-foundation-lemma2a})\\
 &\le& -{\eta\over s_t}(\alpha\lambda_0 \sum_{j=1}^T\Expect(\Vert \bigtriangledown F(x_j)\Vert^2)\\
&&+{\eta\alpha \over s_t}\sum_{j=1}^T\sum_{i=1}^{j-1}(j-i)\beta^{j-i} \left( \Expect(\Vert g_i\Vert^2)\right)+{\alpha L\over 2(1-\beta)^2}\left({\alpha\eta\over s_t}\right)^3{\beta\over 1-\beta}\sum_{j=1}^T\left(\sum_{p=1}^{j-1}\Expect(\Vert g_p\Vert^2)\right)\\
 &\le& -{\eta\over s_t}(\alpha\lambda_0 \sum_{j=1}^T\Expect(\Vert \bigtriangledown F(x_j)\Vert^2)\\
&&+{\eta\alpha \over s_t}\sum_{j=1}^T{\beta\over (1-\beta)^2} \left( \Expect(\Vert g_i\Vert^2)\right)+{\alpha L\over 2(1-\beta)^2}\left({\alpha\eta\over s_t}\right)^3{\beta\over 1-\beta}T\left(\sum_{p=1}^{T}\Expect(\Vert g_p\Vert^2)\right)\\
&&(
by\ 
%\ (3)\ of\ Lemma~\ref{basic-sum-lemma}\ and\ 
Lemma~\ref{basic2-sum-lemma})\\
 &\le& -{\eta\over s_t}(\alpha\lambda_0 \sum_{j=1}^T\Expect(\Vert \bigtriangledown F(x_j)\Vert^2)\\
&&+\left({\eta\alpha\beta \over s_t( 1-\beta)^2} +{\alpha\beta L\over 2(1-\beta)^3}\left({\alpha\eta\over s_t}\right)^3T\right)\sum_{p=1}^{T}(\Expect(\Vert g_p\Vert^2))\\
% &=& -{\eta\over s_t}(\alpha \sum_{j=1}^T\Expect(\Vert \bigtriangledown F(x_j)\Vert^2)\\
%&&+\left({\eta\beta\alpha \over s_t( 1-\beta)^2} +{\alpha\beta L\over 2(1-\beta)^3}\left({\alpha\eta\over s_t}\right)^3T\right)\sum_{p=1}^{T}(\Expect(\Vert g_p\Vert^2))\\
&\le& -{\eta\over s_t}(\alpha\lambda_0 \sum_{j=1}^T\Expect(\Vert \bigtriangledown F(x_j)\Vert^2)+T\left({\alpha\eta\over s_t}\left( {2\sigma_0^2v\over u} \right){\beta\over 1-\beta}\right)\\
&&+\left({\eta\alpha\beta \over s_t( 1-\beta)^2} +{\alpha\beta L\over 2(1-\beta)^3}\left({\alpha\eta\over s_t}\right)^3T\right)\sum_{p=1}^{T}(2\sigma_0^2+ (2+2\sigma_1^2)\Expect(\Vert \bigtriangledown F(x_p)\Vert^2))\\
&\le& -{\eta\over s_t}(\alpha\lambda_0 \sum_{j=1}^T\Expect(\Vert \bigtriangledown F(x_j)\Vert^2)\\
&&+2\sigma_0^2\left({\eta\alpha\beta  \over s_t( 1-\beta)^2} +{\alpha\beta L\over 2(1-\beta)^3}\left({\alpha\eta\over s_t}\right)^3T\right)\\
&&+(2+2\sigma_1^2)\left({\eta\alpha\beta  \over s_t( 1-\beta)^2} +{\alpha\beta L\over 2(1-\beta)^3}\left({\alpha\eta\over s_t}\right)^3T\right)\sum_{p=1}^{T}( \Expect(\Vert \bigtriangledown F(x_p)\Vert^2))\\
&\le& -{\eta\alpha\lambda_0 \over s_t}\sum_{j=1}^T\Expect(\Vert \bigtriangledown F(x_j)\Vert^2)+2\sigma_0^2\left({\eta\alpha\beta  \over s_t( 1-\beta)^2} +{\alpha\beta L\over 2(1-\beta)^3}\left({\alpha\eta\over s_t}\right)^3T\right)\\
&&+\left(\gamma_0\cdot {\eta\alpha \over s_t}\sum_{j=1}^T\Expect(\Vert \bigtriangledown F(x_j)\Vert^2)\right)\ \ (by\ inequality~(\ref{convergence1a-ineqn}))\\
&=& -{(1-\gamma_0)\eta\alpha\lambda_0 \over s_t}\sum_{j=1}^T\Expect(\Vert \bigtriangledown F(x_j)\Vert^2)+2\sigma_0^2\left({\eta\alpha\beta  \over s_t( 1-\beta)^2} +{\alpha\beta L\over 2(1-\beta)^3}\left({\alpha\eta\over s_t}\right)^3T\right).
\end{eqnarray*}

\end{proof}

%***

\begin{lemma}\label{Second1-Basic-lemma}
\begin{eqnarray*}
&&\Expect(\sum_{j=1}^T\Vert x_{j+1}-x_j\Vert^2)	\le {2\sigma_0^2T\over (1-\beta)^2}\left({\alpha\eta\over s_t}\right)^2+ {(2+2\sigma_1^2)\over (1-\beta)^2}\left({\alpha\eta\over s_t}\right)^2\Expect\left(\sum_{i=1}^T ( \Vert \bigtriangledown F(x_i)\Vert^2)\right).  \end{eqnarray*}
	
\end{lemma}

\begin{proof}By Lemma~\ref{basic0-lemma} and Lemma~\ref{basic0b1-lemma}, we have inequalities:
	\begin{eqnarray}
\sum_{j=1}^T\Vert x_{j+1}-x_j\Vert^2)	
\le {1\over (1-\beta)^2}\left({\alpha\eta\over s_t}\right)^2\sum_{i=1}^T (\Vert g_i\Vert^2).
	\end{eqnarray}
	
	Therefore,
	
	\begin{eqnarray*}			&&\Expect\left(\sum_{j=1}^T\Vert x_{j+1}-x_j\Vert^2)	\right)\\		
		&\le& \Expect\left({1\over (1-\beta)^2}\left({\alpha\eta\over s_t}\right)^2\sum_{i=1}^T (\Vert g_i\Vert^2)\right)\le {1\over (1-\beta)^2}\left({\alpha\eta\over s_t}\right)^2 \Expect\left(\sum_{i=1}^T (\Vert g_i\Vert^2)\right)\\
		&\le& {1\over (1-\beta)^2}\left({\alpha\eta\over s_t}\right)^2 \Expect\left(\sum_{i=1}^T (2\sigma_0^2+ (2+2\sigma_1^2)\Vert \bigtriangledown F(x_i)\Vert^2)\right)\ \ \ (by~ Lemma~\ref{First-Basic-lemma})\\
&\le& {1\over (1-\beta)^2}\left({\alpha\eta\over s_t}\right)^2\left( 2\sigma_0^2T+(2+2\sigma_1^2)\Expect\left(\sum_{i=1}^T ( \Vert \bigtriangledown F(x_i)\Vert^2)\right)\right)\\
&\le& {2\sigma_0^2T\over (1-\beta)^2}\left({\alpha\eta\over s_t}\right)^2+ {(2+2\sigma_1^2)\over (1-\beta)^2}\left({\alpha\eta\over s_t}\right)^2\Expect\left(\sum_{i=1}^T ( \Vert \bigtriangledown F(x_i)\Vert^2)\right)	\end{eqnarray*}

\end{proof}

We need the following inequality for convergence

\begin{eqnarray}
    {L\over 2}\cdot{(2+2\sigma_1^2)\over (1-\beta)^2}\left({\alpha\eta\over s_t}\right)^2\le {\gamma_0\eta\alpha \lambda_0\over s_t}. \label{convervence1-second-ineqn}
\end{eqnarray}

We define a few terms for the error upper bound in convergence.

\begin{eqnarray*}
    &&V_1(s_t, T, \sigma_0)={s_t\over (1-2\gamma_0)\eta \alpha\lambda_0 T}(F(x_0)-F(x^*))     \\
    &&V_2(s_t, T, \sigma_0)=\frac{2\sigma_0^2s_t}{\eta\alpha\lambda_0(1-2\gamma_0) }\left({\eta\beta\alpha \over s_t( 1-\beta)^2} +{\alpha\beta L\over 2(1-\beta)^3}\left({\alpha\eta\over s_t}\right)^3T\right) \\
    &&V_3(s_t, T, \sigma_0)=\frac{2s_tL}{\eta\alpha\lambda_0(1-2\gamma_0) }\left({2\sigma_0^2\over (1-\beta)^2}\left({\alpha\eta\over s_t}\right)^2 \right).
\end{eqnarray*}

\begin{theorem}\label{main-thm}
	Suppose $F(.)$ is in $\mathbb{C}_L^1$ and $\inf_x F(x)>-\infty$. Function $G(\xi,x)$ satisfies the conditions in Definition~\ref{condition}. Let $\gamma_0$ be a positive constant that satisfies inequality (\ref{convergence1a-ineqn}).
    % Assume $\beta$, $T$ and $t$ satisfy the inequalities (\ref{beta-inequality}) to (\ref{t-eqn}).  
    Inequalities %(\ref{lambda0-ineqn}), 
    (\ref{convergence1a-ineqn2}) and (\ref{convervence1-second-ineqn}) are satisfied.   Then the algorithm Static-Momentum-GD$(G(.,.), \eta,  x_0, t, T)$ returns  $\min_{1\le i\le T}\Expect(\Vert \bigtriangledown F(x_i)\Vert^2)\le V_1(s_t, T, \sigma_0)+V_2(s_t, T, \sigma_0)+V_3(s_t, T, \sigma_0)$.
\end{theorem}

\begin{proof}
	As $F(x)$ is $L$-Lipschitz smooth, by Lemma~\ref{basic-lemma}, we have 
	\begin{eqnarray*}
		F(x_{j+1})&\le& F(x_j)+(\bigtriangledown F(x_{j}), x_{j+1}-x_j)+{L\over 2}\Vert x_{j+1}-x_j\Vert^2		
	\end{eqnarray*}

Therefore,
	\begin{eqnarray*}
	&&F(x^*)\le F(x_0)+\sum_{j=1}^T\Expect(\bigtriangledown F(x_{j}), x_{j+1}-x_j))+{L\over 2}\sum_{j=1}^T\Expect(\Vert x_{j+1}-x_j\Vert^2)\\		
&\le& F(x_0)-\sum_{j=1}^T\Expect({\eta\over s_t}(\alpha\lambda_0 \langle \bigtriangledown F(x_j), \bigtriangledown F(x_j)\rangle)\\
&&+{L\over 2}\left({2\sigma_0^2T\over (1-\beta)^2}\left({\alpha\eta\over s_t}\right)^2+ {(2+2\sigma_1^2)\over (1-\beta)^2}\left({\alpha\eta\over s_t}\right)^2\Expect\left(\sum_{i=1}^T ( \Vert \bigtriangledown F(x_i)\Vert^2)\right) \right)\\
&&(by\ Lemma~\ref{Second1-Basic-lemma})\\
&\le& F(x_0)+(-{(1-\gamma_0)\eta\alpha\lambda_0 \over s_t}\sum_{j=1}^T\Expect(\Vert \bigtriangledown F(x_j)\Vert^2)+2\sigma_0^2\left({\eta\beta\alpha \over s_t( 1-\beta)^2} +{\alpha\beta L\over 2(1-\beta)^3}\left({\alpha\eta\over s_t}\right)^3T\right)\\
&&+{L\over 2}\left({2\sigma_0^2T\over (1-\beta)^2}\left({\alpha\eta\over s_t}\right)^2+ {(2+2\sigma_1^2)\over (1-\beta)^2}\left({\alpha\eta\over s_t}\right)^2\Expect\left(\sum_{i=1}^T ( \Vert \bigtriangledown F(x_i)\Vert^2)\right) \right)\\
&&(by\ Lemma~\ref{stochastic-foundation-lemma2a})\\
&\le& F(x_0)-{(1-2\gamma_0)\eta\alpha\lambda_0 \over s_t}\sum_{j=1}^T\Expect(\Vert \bigtriangledown F(x_j)\Vert^2)+2\sigma_0^2\left({\eta\beta\alpha \over s_t( 1-\beta)^2} +{\alpha\beta L\over 2(1-\beta)^3}\left({\alpha\eta\over s_t}\right)^3T\right)\\
&&+{L\over 2}\left({2\sigma_0^2T\over (1-\beta)^2}\left({\alpha\eta\over s_t}\right)^2 \right).\ \ \ (by\ inequality~(\ref{convervence1-second-ineqn}))
\end{eqnarray*}

Therefore,
\begin{eqnarray*}
    &&{(1-2\gamma_0)\eta\alpha\lambda_0 \over s_t}\sum_{j=1}^T\Expect(\Vert \bigtriangledown F(x_j)\Vert^2)\\
    &\le& F(x_0)-F(x^*)       +2\sigma_0^2\left({\eta\beta\alpha \over s_t( 1-\beta)^2} +{\alpha\beta L\over 2(1-\beta)^3}\left({\alpha\eta\over s_t}\right)^3T\right)\\
&&+{L\over 2}\left({2\sigma_0^2T\over (1-\beta)^2}\left({\alpha\eta\over s_t}\right)^2 \right).
\end{eqnarray*}

Therefore, we have
\begin{eqnarray*}
    &&\min_{1\le j\le T}\Expect(\Vert \bigtriangledown F(x_j)\Vert^2)\\
    &\le& {s_t\over \eta \alpha\lambda_0(1-2\gamma_0) T}(F(x_0)-F(x^*))   +\frac{2\sigma_0^2 s_t}{\eta\alpha\lambda_0(1-2\gamma_0)}\left({\eta\beta\alpha \over s_t( 1-\beta)^2T} +{\alpha\beta L\over 2(1-\beta)^3}\left({\alpha\eta\over s_t}\right)^3\right)\\
&&+ \frac{2s_tL}{\eta\alpha\lambda_0(1-2\gamma_0)}\left({2\sigma_0^2\over (1-\beta)^2}\left({\alpha\eta\over s_t}\right)^2 \right)\\   
&=& V_1(s_t, T, \sigma_0)+V_2(s_t, T, \sigma_0)+V_3(s_t, T, \sigma_0).
\end{eqnarray*}

\end{proof}

We set parameters for Theorem~\ref{main-thm}.

\begin{lemma}\label{assign-t-st-lemma}
   Assume that $\alpha, \beta, \sigma_1, \eta,\lambda_0$ are positive constants. 
   If inequality~(\ref{convergence1a-ineqn}) holds, then there is an integer constant $c_0$ such that for $t=c_0\ceiling{\log \sqrt{T}}$, and $s=2^t$,
   inequalities (\ref{convergence1a-ineqn2}) and (\ref{convervence1-second-ineqn}) are satisfied for all large $T$. 
\end{lemma}

\begin{proof}
For two positive constants $C_1<C_2$, and another positive constant $C_3$, 
clearly, we have $C_1+\frac{C_3}{x}\le C_2$ if a positive $x$ is selected large enough.  
\end{proof}

\begin{corollary}
	Suppose $F(.)$ is in $\mathbb{C}_L^1$ and $\inf_x F(x)>-\infty$. Function $G(\xi,x)$ satisfies the conditions in Definition~\ref{condition}. Assume that $s_t$ is assigned by  Lemma~\ref{assign-t-st-lemma}.     Then the algorithm Static-Momentum-GD$(G(.,.), \eta,  x_0, t, T)$ returns  $\min_{1\le i\le T}\Expect(\Vert \bigtriangledown F(x_i)\Vert^2)=O({1\over T^{0.5}})$.
\end{corollary}

\begin{proof}
By Lemma~\ref{assign-t-st-lemma}, we have   $s_t=O(T^{0.5})$ and $s_t=\Omega(T^{0.5})$. The convergence rate is around $O({1\over T^{0.5}})$.    This gives the upper bounds:

\begin{eqnarray*}
    V_1(s_t, T, \sigma_0)&=&O\left({1\over T^{0.5}}\right),\\
    V_2(s_t, T, \sigma_0)&=&O\left({1\over T^{0.5}}\right),\\
    V_3(s_t, T, \sigma_0)&=&O\left({1\over T^{0.5}}\right).
\end{eqnarray*}

\end{proof}

%\begin{lemma}\label{assign-t-st-lemma2}
%   Assume that $\alpha, \beta, \sigma_1, \eta,\lambda_0$ are positive constants. 
%   If inequality~\ref{convergence1a-ineqn} holds, then there is an integer constant $c_0$ such that for $t=c_0\ceiling{\delta\log T}$, and $s=2^t$,
%   inequalities (\ref{convergence1a-ineqn}) and (\ref{convervence1-second-ineqn}) are satisfied for all large $T$. 
%\end{lemma}

%\begin{proof}
% It is straighforward to verify them.   
%\end{proof}

%\begin{corollary}
%	Suppose $F(.)$ is in $\mathbb{C}_L^1$ and $\inf_x F(x)>-\infty$. Function $G(\xi,x)$ satisfies the conditions in Definition~\ref{condition} with  $\sigma_0=0$. Assume that $s_t$ is assigned by (\ref{t2a-eqn}).     Then the algorithm GD$(G(.,.), \eta,  x_0, t, T)$ returns  $\min_{1\le i\le T}\Expect(\Vert \bigtriangledown F(x_i)\Vert^2)=O({1\over T^{1-\delta}})$ for any $\delta\in (0,1)$.
%\end{corollary}
%\begin{proof}
%In the case, we have  $\sigma_0=0$ and $s=O(T^{\delta})$. 
%For $\sigma_0=0$, we have $V_2(s_t, T, \sigma_0)=V_3(s_t, T, \sigma_0)=0$, and $V_1(s_t, T, \sigma_0)=O({1\over T^{1-\delta}})$. The convergence rate is  $O({1\over T^{1-\delta}})$.    
%\end{proof}

%\input{parallel-part.tex}

%\documentclass[10pt]{article}

%\usepackage[margin=0.75in]{geometry}
%\usepackage{amsmath}
%\usepackage{graphicx}
%\usepackage{float}
%\usepackage{caption}
%\usepackage{microtype}

%\begin{document}

\section{Simulation Study: AdamW vs.\ Arithmetically-Simple-AdamW}
We evaluated the proposed Arithmetically-Simple-AdamW  optimizer against the baseline AdamW ~\cite{WardWuBottou19,XieWuWard2020}
%~\cite{li2026frac} 
 formulation used in our implementation.
The main goal of the simulation was to compare the two methods in terms of both optimization performance and computational speed.
For optimization performance, we examined how well each method minimized the same synthetic linear-regression problems.
For computational performance, we focused on the execution time of the optimizer update.

\subsection{Arithmetically-Simple-AdamW  Method and Implementation}

The main idea behind Arithmetically-Simple-AdamW  is to replace the square-root and division operations in AdamW's adaptive parameter update with power-of-two scaling.
In AdamW, the first moment is divided by the square root of the second moment plus epsilon, whereas Arithmetically-Simple-AdamW  approximates this scaling factor using a power of two.
For each parameter and iteration, Arithmetically-Simple-AdamW  uses the exponent-selection rule $t=\left\lfloor(q+1)/2\right\rfloor$, where $q$ is the binary exponent of the second moment plus epsilon when its significand is normalized to the interval $[1,2)$.
The approximate denominator is two raised to the selected exponent $t$.
The resulting power-of-two representation allows the adaptive scaling to be implemented through direct manipulation of the IEEE-754 floating-point exponent.
For analytic simplicity, the bias term was omitted for both AdamW and Arithmetically-Simple-AdamW .
An important part of Arithmetically-Simple-AdamW  is how the power-of-two scaling is implemented.
Our initial high-level PyTorch implementation required several separate tensor operations for exponent extraction and adaptive scaling.
To reduce this overhead, we implemented the adaptive-scaling operation as a native C++ extension.
The final version uses branchless IEEE-754 exponent manipulation and is compiled using \texttt{-O3 -march=native}.
Instead of explicitly constructing the power-of-two denominator and performing a conventional floating-point division, the implementation directly adjusts the exponent bits of the first-moment value.
This specialized operation assumes that the first-moment values and scaled results remain in the normal float32 range, with positive, finite, normal values of the second moment plus epsilon.
The loop was also written so that the compiler could automatically vectorize the calculation.
This reduces intermediate tensor operations and allows multiple float32 values to be processed through single-instruction, multiple-data (SIMD) instructions.
The first- and second-moment calculations and the parameter update remain in PyTorch; only the adaptive-scaling operation is performed by the specialized native routine.

\subsection{Linear-Regression Dataset and Experimental Setup}

Our synthetic linear-regression benchmark was motivated by the Gaussian-data experiment in Section~5.1 of Ward et al.~\cite{WardWuBottou19,XieWuWard2020}.
%~\cite{ward2020adagrad}.
%
For the optimization experiment, we generated 20 independent synthetic datasets using random seeds 1 through 20.
Each dataset contained 1,000 observations and 2,000 parameters.
The design matrix and the ground-truth parameter vector were sampled from standard normal distributions, and the target vector was generated directly from the linear model.
Both optimizers received the same dataset and the same zero initialization for each seed.
The loss function was one-half of the mean squared prediction error between the model predictions and the target values~\cite{WardWuBottou19,XieWuWard2020}.
%~\cite{ward2020adagrad}.
%
The mean averages the squared errors over the observations, while the factor of one-half simplifies the gradient expression without changing the minimizers of the least-squares objective.
We used the entire dataset to calculate the loss and gradient at every iteration.
Thus, our experiments adapt the paper's Gaussian least-squares benchmark to a full-batch comparison.
Both methods were run for exactly 5,000 optimization iterations on each dataset using the same hyperparameters: a learning rate of 0.001, first-moment coefficient of 0.9, second-moment coefficient of 0.999, weight decay of 0.01, and epsilon of $10^{-8}$.

\subsection{Optimization Experiment}

\begin{figure}[H]
    \centering
    \includegraphics[width=0.92\linewidth]{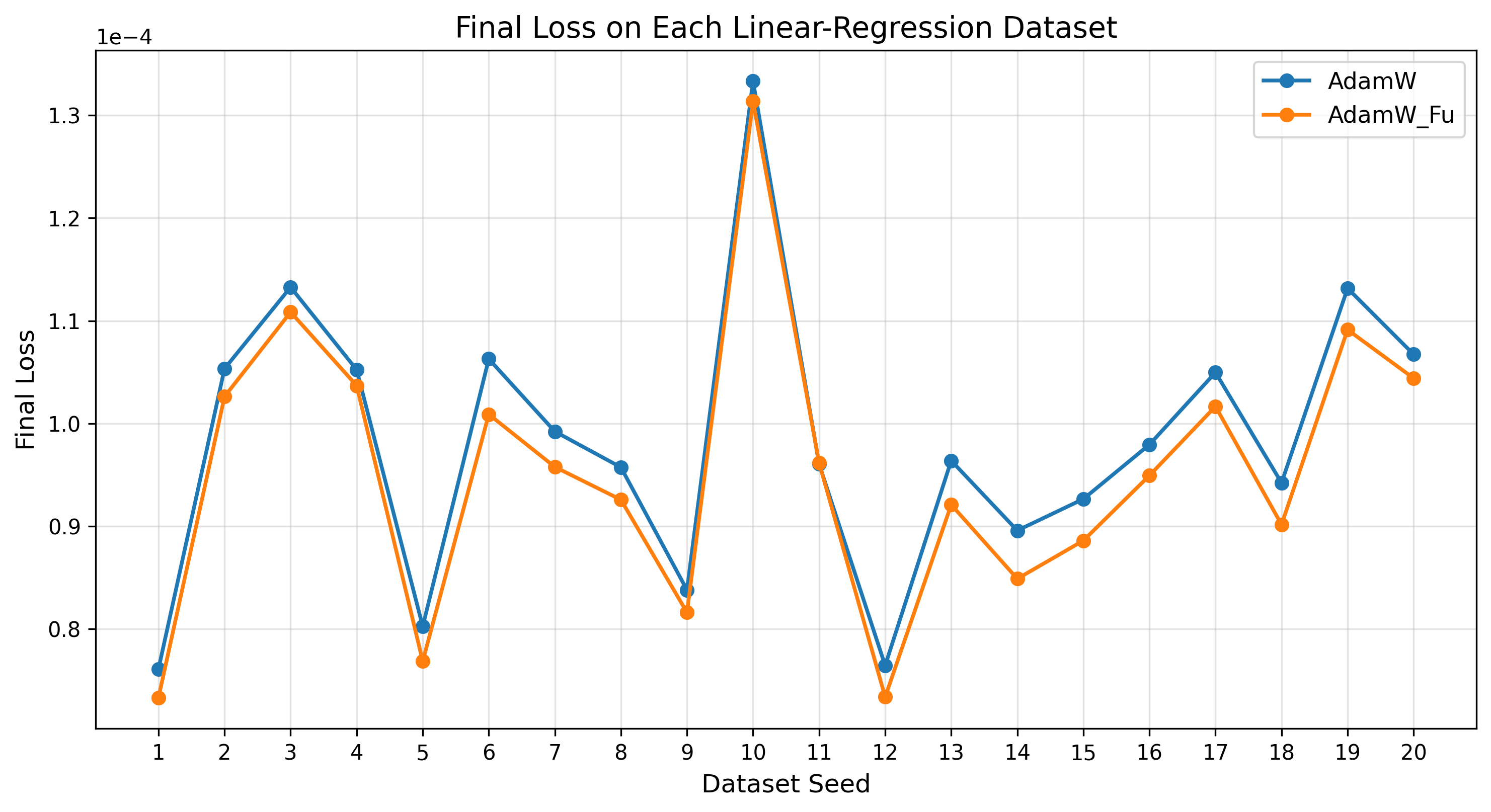}
    \caption{
    Final loss after 5,000 optimization iterations on each of the 20 independently generated synthetic linear-regression datasets.
    Arithmetically-Simple-AdamW  obtained the lower final loss on 19 of the 20 datasets.
    }
    \label{fig:final_loss}
\end{figure}

AdamW obtained a mean final loss of $9.8340\times10^{-5}$ with a standard deviation of $1.3751\times10^{-5}$.
Arithmetically-Simple-AdamW  obtained a lower mean final loss of $9.5254\times10^{-5}$ with a standard deviation of $1.3955\times10^{-5}$.
For each dataset, we calculated the difference between the Arithmetically-Simple-AdamW  and AdamW final losses as a percentage of the AdamW final loss.
The mean of these paired percentage differences was approximately $-3.21\%$, where a negative value means that Arithmetically-Simple-AdamW  obtained the lower loss.
As shown in Figure~\ref{fig:final_loss}, Arithmetically-Simple-AdamW  produced the lower final loss on 19 of the 20 datasets.
The only exception was dataset seed 11, where Arithmetically-Simple-AdamW  produced a final loss approximately 0.13\% higher than AdamW.
Overall, the results show an average improvement, with lower final loss on most of the tested problems.

For both methods, the best recorded loss occurred at the final 5,000th iteration across the tested datasets.
These results therefore describe a comparison under the same fixed optimization budget.

\subsection{Optimizer-Step Timing Experiment}

\begin{figure}[H]
    \centering
    \includegraphics[width=0.82\linewidth]{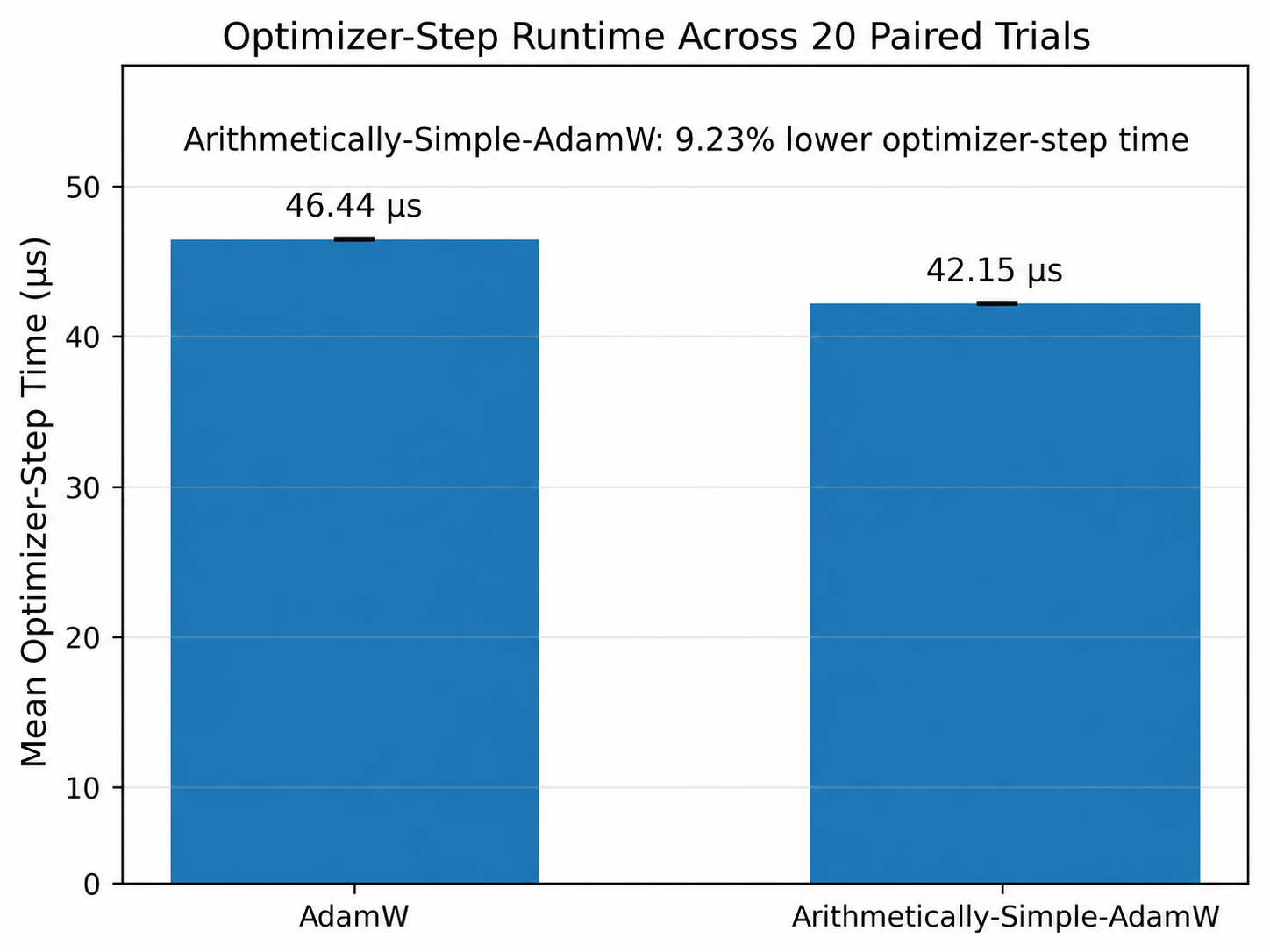}
    \caption{
    Mean optimizer-step execution time over 20 paired timing trials, with 5,000 updates per optimizer in each trial.
    Error bars show the standard deviation of the trial-average step times.
    Arithmetically-Simple-AdamW  reduced optimizer-step time by approximately 9.23\% relative to the baseline implementation.
    }
    \label{fig:runtime}
\end{figure}

The computational-speed experiment was performed separately from the 20-dataset optimization experiment.
It used one fixed linear-regression problem for 20 paired timing trials, with each optimizer completing 5,000 measured updates per trial.
Before the measured trials, each optimizer completed 100 warm-up iterations using separate parameter vectors.
Each measured run restarted from the same zero initialization.
The two optimizers were executed sequentially, with their order alternated between trials to reduce execution-order bias.
The timing experiments were performed on an AMD Ryzen Threadripper PRO 5955WX CPU using float32 tensors.
The main timing measurement focused specifically on \texttt{optimizer.step()}, which includes the moment calculations, adaptive scaling, decoupled weight decay, and parameter update.
We focus on this measurement because it captures the update cost directly affected by Arithmetically-Simple-AdamW , excluding objective evaluation and gradient computation, which use the same code for both methods.
For each trial, we averaged the measured times over the 5,000 optimizer steps and then summarized these trial averages across the 20 repetitions.
The mean optimizer-step time for AdamW was approximately $46.44~\mu\mathrm{s}$ (microseconds).
The corresponding mean for Arithmetically-Simple-AdamW  was approximately $42.15~\mu\mathrm{s}$.
The mean paired reduction was approximately 9.23\%, corresponding to a saving of about $4.29~\mu\mathrm{s}$ per update.
Arithmetically-Simple-AdamW  was faster in all 20 paired timing trials.
The standard deviation of the paired percentage differences was approximately 0.20 percentage points, indicating that the advantage was consistent across these repetitions.
Figure~\ref{fig:runtime} summarizes the optimizer-step timing results.
%

% ============================================================
% FIGURE 2: OPTIMIZER-STEP RUNTIME
% ============================================================

\subsection{Summary}

The simulations show that Arithmetically-Simple-AdamW  achieved a lower optimizer-update cost and improved optimization performance in the tested setting.
Across the 20 independently generated linear-regression datasets, its mean paired final-loss reduction was approximately 3.21\%, with lower final loss on 19 datasets.
The optimized implementation also reduced mean optimizer-step time by approximately 9.23\% relative to our PyTorch AdamW baseline.
These results support the use of power-of-two adaptive scaling together with a specialized native implementation for this workload.
The findings are limited to the tested CPU, float32 implementation, synthetic linear regression problems, and hyperparameters.
Further experiments on other learning tasks and hardware would be needed to determine how broadly these improvements extend.
%
%\bibliographystyle{plain}
%\bibliography{refs}

%\end{document}

%\section{Conclusions}
%In this paper we give the convergence analysis about static Adam, and show that it can be smoothly integrated into the parallel frame work developed by Fu. The static Adam is arithmetically simple that each division of the form $x/2^t$ with integer $t$. A future research topic may improve the model so that it has communication among the processors to speed up the gradient descent.

\bibliographystyle{abbrv}
\bibliography{bib}

\begin{thebibliography}{10}

\bibitem{DuchiHazanSinger2011}
J.~Duchi, E.~Hazan, and Y.~Singer.
\newblock Adaptive subgradient methods for online learning and stochastic optimization.
\newblock {\em The Journal of Machine Learning Research}, 12:2121–2159, 2011.

\bibitem{Fu2026Adaptivity}
B.~Fu.
\newblock Adaptivity via a parallel architecture for stochastic gradient methods.
\newblock {\em arXiv preprint arXiv:2607.28902}, 2026.

\bibitem{KingmaBa2015}
D.~P. Kingma and J.~Ba.
\newblock Adam: A method for stochastic optimization.
\newblock arXiv:1412.6980, 2015.

\bibitem{Loshchilov2019Decoupled}
I.~Loshchilov and F.~Hutter.
\newblock Decoupled weight decay regularization.
\newblock In {\em International Conference on Learning Representations (ICLR)}, 2019.

\bibitem{BottouCurtisNocedal2018}
J.~N. Léon~Bottou, Frank E.~Curtis.
\newblock Optimization methods for large-scale machine learning.
\newblock {\em SIAM Reviews}, 60(2):223–311, 2018.

\bibitem{DBLP:journals/corr/abs-1002-4908}
H.~B. McMahan and M.~J. Streeter.
\newblock Adaptive bound optimization for online convex optimization.
\newblock {\em CoRR}, abs/1002.4908, 2010.

\bibitem{NemirovskiJuditskyLanShapiro2009}
A.~Nemirovski, A.~Juditsky, G.~Lan, and A.~Shapiro.
\newblock Robust stochastic approximation approach to stochastic programming.
\newblock {\em SIAM Journal on Optimization}, 19:1574--1609, 2009.

\bibitem{WardWuBottou19}
R.~Ward, X.~Wu, and L.~Bottou.
\newblock Adagrad stepsizes: Sharp convergence over nonconvex landscapes.
\newblock {\em Journal of Machine Learning Research}, 21 (219):1--30, 2020.

\bibitem{XieWuWard2020}
Y.~Xie, X.~Wu, and R.~Ward.
\newblock Linear convergence of adaptive stochastic gradient descent.
\newblock In {\em Proceedings of the Twenty Third International Conference on Artificial Intelligence and Statistics}, pages PMLR 108:1475--1485, 2020.

\bibitem{DBLP:journals/corr/abs-1212-5701}
M.~D. Zeiler.
\newblock {ADADELTA:} an adaptive learning rate method.
\newblock {\em CoRR}, abs/1212.5701, 2012.

\end{thebibliography}

\end{document}